\documentclass[12pt]{article}

\usepackage{newtxtext,newtxmath}

\usepackage{graphicx}
\graphicspath{}

\usepackage[letterpaper,margin=1in]{geometry}

\renewenvironment{abstract}
	{\quotation}
	{\endquotation}

\date{}

\makeatletter
\renewcommand{\fnum@figure}{\textbf{Figure \thefigure}}
\renewcommand{\fnum@table}{\textbf{Table \thetable}}
\makeatother

\usepackage{scicite}

\usepackage{url}

\def\scititle{
	Breaking speed scaling in quadrupedal robots via Huygens’ coupled-pendulum dynamics
}
\title{\bfseries \boldmath \scititle}

\author{
	Yucheng Tao$^{1\dagger}$,
	Yongbin Jin$^{1,2,3\ast\dagger}$,
	Shaowen Cheng$^{2}$,\and
	Xianwei Liu$^{3}$,
	Yanyan Yuan$^{3}$,
	Yanhong Liang$^{1}$,\and
	Chengkai Su$^{1}$,
	Chaojie Fu$^{1}$,
	Guorong Lan$^{3}$,\and
	Wei Yang$^{1\ast}$,
	Hongtao Wang$^{1,2,3\ast}$\and
	\small$^{1}$Center for X-Mechanics, Zhejiang University, Hangzhou, China.\and
	\small$^{2}$ZJU-Hangzhou Global Scientific and Technological Innovation Center, Hangzhou, China.\and
	\small$^{3}$Mirrorme Technology Co., Ltd., Shanghai, China.\and
	\small$^\ast$Corresponding author. Email: yongbinjin@zju.edu.cn\and
	\small$^\dagger$These authors contributed equally to this work.
}

\begin{document} 

\maketitle

\begin{abstract} \bfseries \boldmath
Achieving biological-level running speeds has largely been pursued through advances in control algorithms, which improve the utilization of existing hardware. However, the ultimate speed limits remain governed by the underlying force and torque requirements of rapid locomotion, which are typically addressed through increased actuator capacity. Inspired by Huygens' coupled pendulums, we demonstrate that superior locomotion can emerge from principled exploitation of intrinsic dynamics rather than brute-force hardware scaling. Inter-limb inertial coupling redistributes energy across the gait cycle and reduces peak joint torque required for rapid periodic motion, thereby expanding the achievable speed without proportional increases in actuator capability. Incorporating hardware parameters as additional design variables further extends this analysis into a co-optimization framework, enabling the systematic utilization of inertial coupling in robot design. Guided by this framework, a quadruped robot achieves a running speed of 10.74 m/s (Froude number 21.4) and completes a 100-meter sprint in 12.2 seconds, representing the first legged robot to surpass 10 m/s. These results establish inertial coupling as an underlying mechanism governing high-speed legged locomotion and highlight its role in reducing force requirements, offering new insights into the design of agile robotic systems.
\end{abstract}

\subsection*{INTRODUCTION}
\noindent

\begin{figure}[!htbp]
	\centering
	\includegraphics[width=\textwidth,height=0.58\textheight,keepaspectratio]{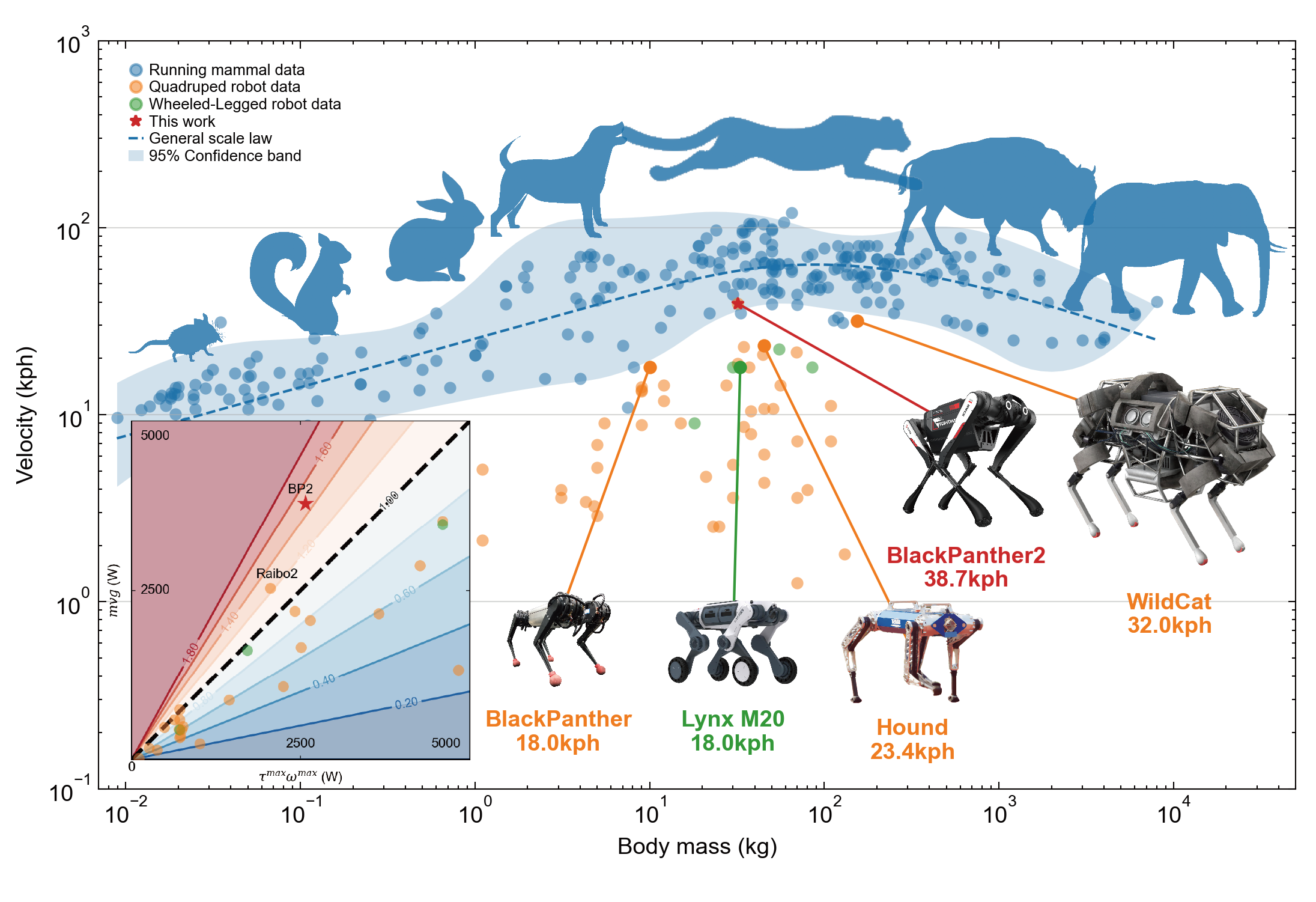}

	\caption{\textbf{Statistics of the maximum speed and body mass of mammals and quadrupedal robots in logarithmic scales.}
		There are total 171 mammalian species of mammals with size ranging from an adult mouse to an elephant. Data for mammalian species are obtained from Hirt et al. \cite{hirt_general_2017}, in which body mass and running speed adhere to a general scaling law: $v=25.5m^{0.26}(1-e^{-22m^{-0.6}})$. The lower-left inset summarizes peak locomotion performance and actuator capability for representative quadrupedal robots. Detailed data are provided in Table~\ref{tab:robot_specs}.}
	\label{fig:statistics}
\end{figure}

A central goal of legged robotics is to replicate and ultimately surpass biological locomotion, with maximum running speed serving as a key integrative performance metric. Despite substantial progress, the maximum running speed of current robotic systems, even for wheel-legged robots, remains significantly lower than that of biological organisms (Fig.~\ref{fig:statistics}). In nature, high-speed locomotion is critical for survival, emerging from the evolutionary pressures of pursuit and escape \cite{alexander2003principles,dickinson2000animals}. Even in humans, where natural selection pressures have been largely diminished, the pursuit of speed persists through athletic competition. However, improvements in sprint performance have proven remarkably difficult: it took more than seven decades after the revival of the modern Olympic Games to break the first 10-second barrier, highlighting the challenge of pushing speed limits. Importantly, such performance gains are not achieved through increased power alone, but arise from a coordinated interplay among morphology, musculoskeletal elasticity, and intersegmental dynamics and gait \cite{weyandFasterTopRunning2000,morinTechnicalAbilityForce2011}. This raises a fundamental question: can a deeper understanding and exploitation of intrinsic dynamics reduce the torque requirements of high-speed locomotion, thereby enabling performance beyond conventional actuator-limited scaling?

The aspiration for high-speed locomotion in robotics is driven not only by benchmark comparisons, but also by the broader need to expand the functional envelope of legged systems for tasks requiring rapid response and agile traversal of complex environments. Yet high-speed locomotion remains intrinsically challenging, as it compresses stability margins and leads to a sharp increase in energy dissipation. Small mismatches in state estimation and contact timing can rapidly accumulate, causing instability or excessive energy cost that suppresses performance. These challenges reflect the increasing difficulty of regulating forces and motions in systems whose underlying dynamics become progressively more demanding at higher speeds.

Over the past few years, deep reinforcement learning (RL) has fundamentally advanced robotic locomotion by improving the ability and timeliness to handle complex dynamics and high-dimensional control. Techniques such as asymmetric actor–critic \cite{nahrendra_dreamwaq_2023,long_hybrid_2024} and teacher–student learning \cite{choi_learning_2023,lee_learning_2020,kumar_rma_2021} can be interpreted as mechanisms for enhancing state estimation under partial observability, enabling more accurate inference of system states that are critical for agile maneuvers and parkour over terrains \cite{he_attention-based_2025,cheng_extreme_nodate,hoeller_anymal_2024,luo_pie_2024,wang_sf-tim_2024,kim_high-speed_2025}.
In the domain of high-speed locomotion, a two-stage training framework has been proposed to align control with system-specific dynamics for high-speed locomotion, achieving substantial performance gains \cite{jin_high-speed_2022}. These advances extend locomotion capabilities to increasingly dynamic regimes, with representative systems achieving speeds of 4–6 m/s \cite{crowley_optimizing_2023,Unitree_H1_Unitree_Robotics,Unitree_A2_Unitree_Robotics,Unitree_B2_Unitree_Robotics,X30_DEEP_Robotics,Lynx_M20_DEEP_Robotics}. 

For the ultimate speed, the limitations of this paradigm become more pronounced, particularly in the discrepancy between simulation and real-world performance \cite{mishraHACLHistoryAwareCurriculum2025}.
Policies that achieve high speeds in simulation often degrade significantly on hardware due to unmodeled dynamics and contact inconsistencies. To address this gap, several works focus on simulation fidelity. Hwangbo et al. introduce learned actuator models to capture motor dynamics beyond idealized assumptions, breaking the previous speed record of ANYmal by 25\% \cite{hwangboLearningAgileDynamic2019}. Shin et al. incorporate a quasi-direct drive (QDD) motor model into RL, enabling the Hound robot to achieve a substantial 44\% increase in speed. Similarly, Miller et al. apply system identification to Spot to align simulation parameters with real-world behavior, resulting in improved high-speed performance from 3.8 m/s to 5.2 m/s \cite{miller_high-performance_2025}.
Collectively, these studies highlight that while learning-based controllers have become increasingly powerful, their effectiveness at high speeds is ultimately bounded by the fidelity of the underlying dynamics. 

Despite rapid progress in learning-based control, the achievable performance of legged robots remains fundamentally bounded by their physical design. As shown in the inset of Fig.~\ref{fig:statistics}, the peak locomotion performance of most quadrupedal robots scales approximately with actuator capability, characterized by the ratio between the robot’s peak mechanical output ($mgv$) and the maximum motor power envelope. Across existing platforms, this dimensionless ratio is typically distributed between 0.6 and 0.8, suggesting a common hardware-dependent performance boundary. The ratio definition of this dimensionless parameter is similar to that of total cost of transport (TCOT), but it focuses more on the ultimate performance. Notably, although modern robotic actuators already surpass biological muscles in both power density and peak torque \cite{wensingProprioceptiveActuatorDesign2017}, these hardware advantages do not yet translate into a proportional lead in locomotion speed. This persistent performance gap underscores that further improvements cannot rely solely on brute-force scaling of actuation. Interestingly, Raibo2 \cite{hwangbo_raibo2_2025}, currently the only quadrupedal robot to complete a full marathon, reaches a ratio of approximately 1.2 through extreme energy optimization, indicating that locomotion performance need not scale solely through increases in actuator capacity.

Existing efforts in hardware design explore biologically inspired topological augmentations, such as articulated spines, tails, and specialized feet, which can enhance energy efficiency, stability, or acceleration \cite{bhattacharya_learning_2019,norby_enabling_2021,heim_designing_2016}. Beyond topology, a more fundamental line of inquiry concerns how dynamical understanding can inform hardware design. Over the past decades, simplified models such as the Spring-Loaded Inverted Pendulum (SLIP) have provided valuable insights into the relationships between morphology, stiffness, and achievable speed \cite{blickhan_spring-mass_1989,dario_how_1993}. These analyses clarify that locomotion performance is governed not only by actuator capacity, but also by how mechanical properties shape energy storage, transfer, and release throughout the gait cycle.

To date, most existing quadrupedal robots continue to follow a design paradigm inherited from early electrically actuated platforms, in which minimizing leg inertia is considered essential for achieving accurate locomotion control \cite{park_mit_2014}. Lightweight limbs are therefore favored to reduce effective inertia, improve responsiveness, and mitigate impact forces during foot–ground contact \cite{wensingProprioceptiveActuatorDesign2017}.
This preference is also reinforced by the limitations of simulation: because impact dynamics remain among the least accurately modeled aspects of legged locomotion, reducing limb inertia can decrease sensitivity to contact-modeling errors and alleviate sim-to-real discrepancies. This may partly explain why learning-based controllers often achieve more reliable transfer on robots with lighter legs.
However, this design paradigm introduces an inherent trade-off between force generation, structural strength, and system mass. Achieving higher torque typically entails increased motor inertia, while the large impact forces associated with high-speed locomotion demand greater structural strength, both of which contribute to increased limb mass. Interestingly, biological systems do not strictly adhere to a lightweight design principle. Elite high-speed animals, such as cheetahs and greyhounds, possess substantial limb mass, with the legs accounting for a significant fraction of total body mass \cite{hudson_functional_2011}. Rather than minimizing inertia, these systems exploit coordinated whole-body dynamics to achieve rapid and efficient locomotion. This discrepancy suggests that current hardware design strategies may overlook important dynamical mechanisms, limiting the ability of robots to fully exploit their physical potential.

\begin{figure}[!htbp]
	\centering
	\includegraphics[width=\textwidth,height=0.58\textheight,keepaspectratio]{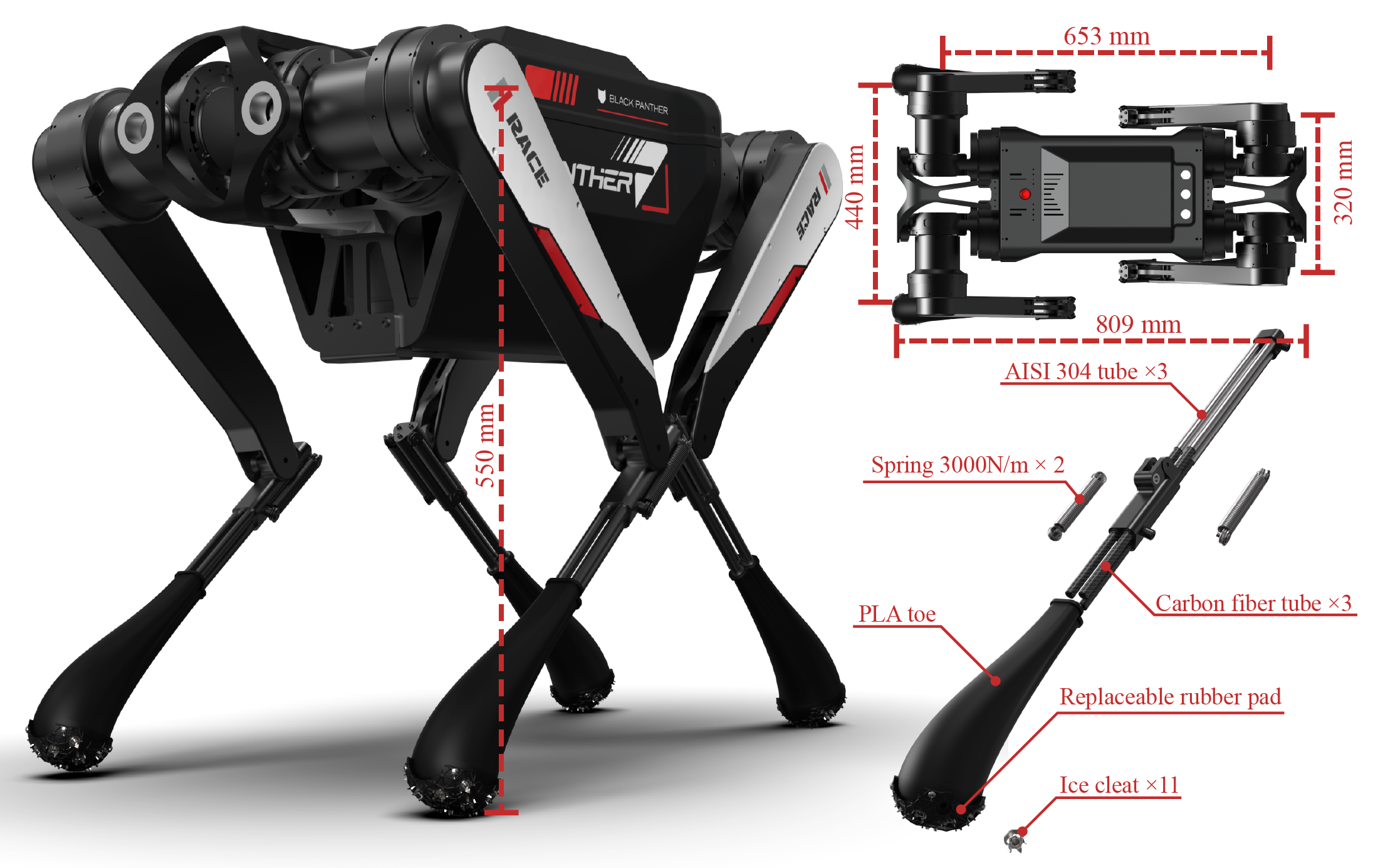}
	\caption{\textbf{Mechanical structure diagram of BP2.} BP2 has a standing height of 55 cm, a body length of 80 cm, a width of 44 cm, and a total mass of 36 kg with battery. The shank structure is enlarged to clarify the assembly of the spring components and foot.}
	\label{fig:bp2}
\end{figure}

Departing from the prevailing intuition that limb inertia should be minimized, we instead investigate how inertia can be actively exploited. In 1665, Christiaan Huygens observed that two pendulum clocks suspended from a shared beam spontaneously synchronized through weak mechanical coupling. Inspired by this phenomenon \cite{tian_measure_2019}, we construct and analyze a simplified coupled pendulum system, showing that interactions between pendulums can redistribute energy and significantly reduce the peak actuation required by individual components.

Building on this insight, we formulate a hybrid dynamics model for quadruped robots and perform trajectory optimization to probe the mechanics of high-speed running. The results reveal that analogous coupling effects naturally emerge in quadrupedal running, where it systematically reduces the peak joint torque required to sustain rapid periodic motion. This finding identifies inertial coupling as a key mechanism underlying efficient high-speed locomotion.
By further incorporating hardware parameters into the optimization, we establish a hardware–trajectory co-optimization framework that directly links hardware parameters to achievable performance, providing quantitative guidance for high-speed robot design. The complete workflow and main results of this study are summarized in Movie~S1.

Guided by this approach, we design and build a high-speed quadruped robot, BP2 (Fig.~\ref{fig:bp2}), and validate the proposed framework through real-world experiments. In outdoor trials, BP2 achieves a maximum running speed of 10.74 m/s, corresponding to a Froude number of 21.4. In indoor treadmill experiments, BP2 reaches a peak running speed of 13.2 m/s. It further completes a 100-meter sprint in 12.2 s, meeting the National Level II Athlete standard in China (Table~\ref{tab:100m_standard}). To the best of our knowledge, this represents the fastest recorded speed for a quadrupedal robot to date.

\subsection*{RESULTS}
\subsubsection*{Huygens' coupled pendulum}
\begin{figure}[!htbp]
	\centering
	\includegraphics[width=\textwidth,height=0.58\textheight,keepaspectratio]{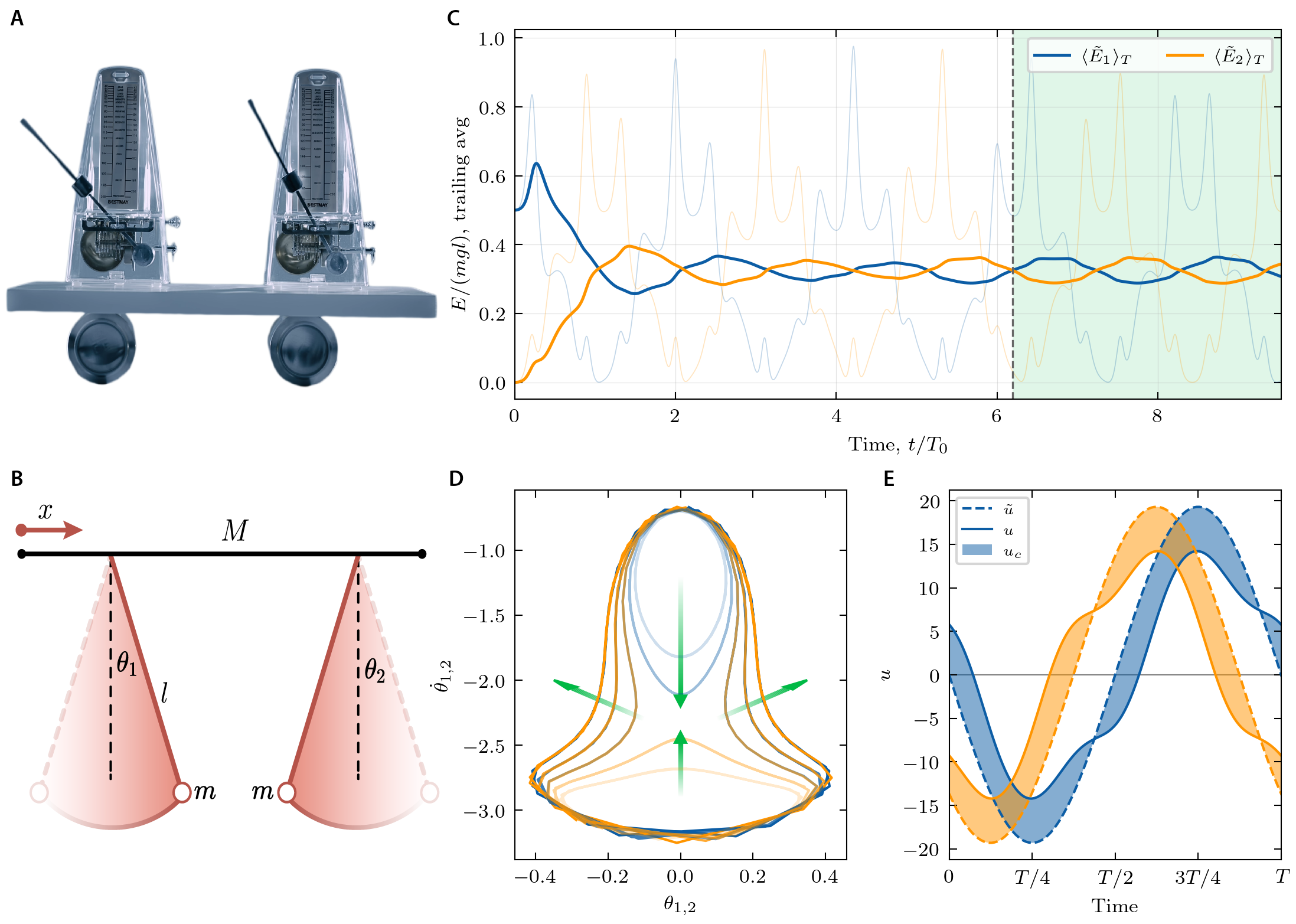}
	\caption{\textbf{Analysis of the coupled dynamic model.}
    (\textbf{A}) Experimental setup for Huygens' coupled pendulums.
    (\textbf{B}) The simplified coupled dynamic model.
	(\textbf{C}) Dimensionless per-pendulum energies $\tilde{E}_i = E_i/(mgl)$ vs.\ normalized time $t/T_0$ (with $T_0 = 2\pi\sqrt{l/g}$). Light traces are instantaneous, bold traces are trailing time-averages $\langle \tilde{E}_{1,2} \rangle_T$.
    (\textbf{D}) Poincaré section analysis of the coupled dynamic system.
    (\textbf{E}) Active torque requirements under specific kinematic conditions ($R = 1,\,A = 1\,\text{rad},\,\omega = 5\,\text{Hz},\,\phi = \tfrac{1}{4}\pi$).}
	\label{fig:model_analysis}
\end{figure}

The high dimensionality and hybrid nonlinear dynamics, arising from intermittent ground contact and strongly coupled limb interactions, make a full quadrupedal system difficult to analyze in terms of isolated dynamic effects. Therefore, to investigate the underlying coupling mechanism, we construct a simplified dynamics model analogous to Huygens' pendulums (Fig.~\ref{fig:model_analysis}A). In this model, the coupling strength is defined by the beam-to-pendulum mass ratio, $R=M/m$ (Fig.~\ref{fig:model_analysis}B). Dynamic simulations indicate that this ratio is a key parameter governing both energy exchange and synchronization \cite{tian_measure_2019}. Specifically, a lower ratio $R$,corresponding to a heavier pendulum, promotes synchronization between the two pendulums, even when they are initialized with different energy states. The resulting passive redistribution is visualized from two complementary viewpoints. In the time domain (Fig.~\ref{fig:model_analysis}C), pendulum~1 is released from $A_0 = 60^{\circ}$ while pendulum~2 starts at rest; the trailing time-averaged dimensionless energies $\langle \tilde{E}_{1,2}\rangle_T$ converge from $(0.5,0)$ to near-equipartition $\langle \tilde{E}_{1,2}\rangle_T \approx 0.25$ within roughly six natural periods (shaded band), despite the persistent oscillation of the instantaneous energies (light traces). The Poincaré section (Fig.~\ref{fig:model_analysis}D) confirms this trend by sweeping the coupling ratio $R$ in passive free-oscillation simulations: as $R$ decreases (from faded to solid traces), the phase-space sections of the two pendulums progressively contract from independent orbits toward a single, coincident path, with no external actuation involved. The underlying mechanism is a spontaneous energy transfer mediated by the off-diagonal terms of the mass matrix: as $R$ decreases, the inertial coupling through the shared movable base becomes dominant, breaking the independence of the two pendulums and enabling bidirectional energy exchange. This passive entrainment provides a dynamical foundation for the active control strategies.

Analogous to motor actuation, we extend the passive dynamics model by incorporating active torques. The resulting dimensionless active torque for each pendulum is derived as:

\begin{equation}
	u_{i}=\tilde{u}_{i}+\underbrace{\frac{C_{i}}{R+2} \sum_{j=1}^{2}\left(S_{j} \theta_{j}^{\prime 2}-C_{j} \theta_{j}^{\prime \prime}\right)}_{u_{i}^{c}}
	\label{eq:active torque} 
\end{equation}

\noindent
where $\tilde{u}_{i}$ represents the dimensionless active torque requirement in the absence of coupling (with the beam fixed), $u_{i}^{c}$ denotes the change in torque demand induced by coupling effects, and $S_{i}$ and $C_{i}$ are shorthand notations for $\sin \theta_{i}$ and $\cos \theta_{i}$, respectively, a convention adhered to throughout this paper. This expression explicitly illustrates the potential reduction in the active torque requirement by leveraging the dynamic coupling between the pendulum and the beam. The complete model formulation and the detailed derivation are provided in Methods.

By substituting appropriate values of $R$ into Eq.~\ref{eq:active torque}
under prescribed kinematic conditions where $\theta_{1} = A \sin(2\pi \omega t)$ and
$\theta_{2} = A \sin(2\pi \omega t + \phi)$,  the active torque over a complete motion cycle can be evaluated in terms of both magnitude and composition.
Fig.~\ref{fig:model_analysis}E presents the resulting joint torques with coupling effects explicitly included, where the shaded regions denote the contribution arising from coupling. Notably, this coupling significantly reduces the peak torque demand, which is critical for actuator limitations.

\subsubsection*{Inertial coupling in robot running}

Analogously, a quadrupedal robot can be viewed, to a first approximation, as a set of limbs dynamically coupled through a shared body, where each leg behaves as an oscillatory subsystem interacting with others via the trunk. This perspective motivates the hypothesis that, similar to Huygens' coupled pendulums, properly structured inertial coupling within the robot may actively contribute to high-speed locomotion rather than serving solely as a passive load.

To quantitatively analyze this effect, we formulate a hybrid dynamical model that captures both continuous running dynamics and discrete contact events. In the following analysis, we focus on the continuous phase of the dynamics, which determines how inertial coupling redistributes energy and modulates the actuation demand during periodic motion. The corresponding equations of motion are

\begin{equation}
\mathbf{M}(\mathbf{q}) \ddot{\mathbf{q}}
+\mathbf{C}(\mathbf{q}, \dot{\mathbf{q}})
+\mathbf{G}(\mathbf{q})
=
\boldsymbol{\tau}_{a}
+\boldsymbol{\tau}_{p}
+\mathbf{J}^{T}\mathbf{F}.
\label{eq:continuous_dynamics}
\end{equation}

\noindent
here, $\mathbf{q}$ denotes the generalized coordinates; $\mathbf{M}(\mathbf{q})$ is the mass matrix; $\mathbf{C}(\mathbf{q},\dot{\mathbf{q}})$ contains Coriolis and centrifugal effects; and $\mathbf{G}(\mathbf{q})$ represents gravity. The terms $\boldsymbol{\tau}_{a}$ and $\boldsymbol{\tau}_{p}$ denote active actuation and passive torques, respectively, and $\mathbf{J}^{T}\mathbf{F}$ represents the generalized force induced by ground contact. The complete hybrid formulation, including the impact map at discrete contact events, is provided in Materials and Methods.

\begin{figure}[!htbp]
	\centering
	\includegraphics[width=\textwidth,height=0.85\textheight,keepaspectratio]{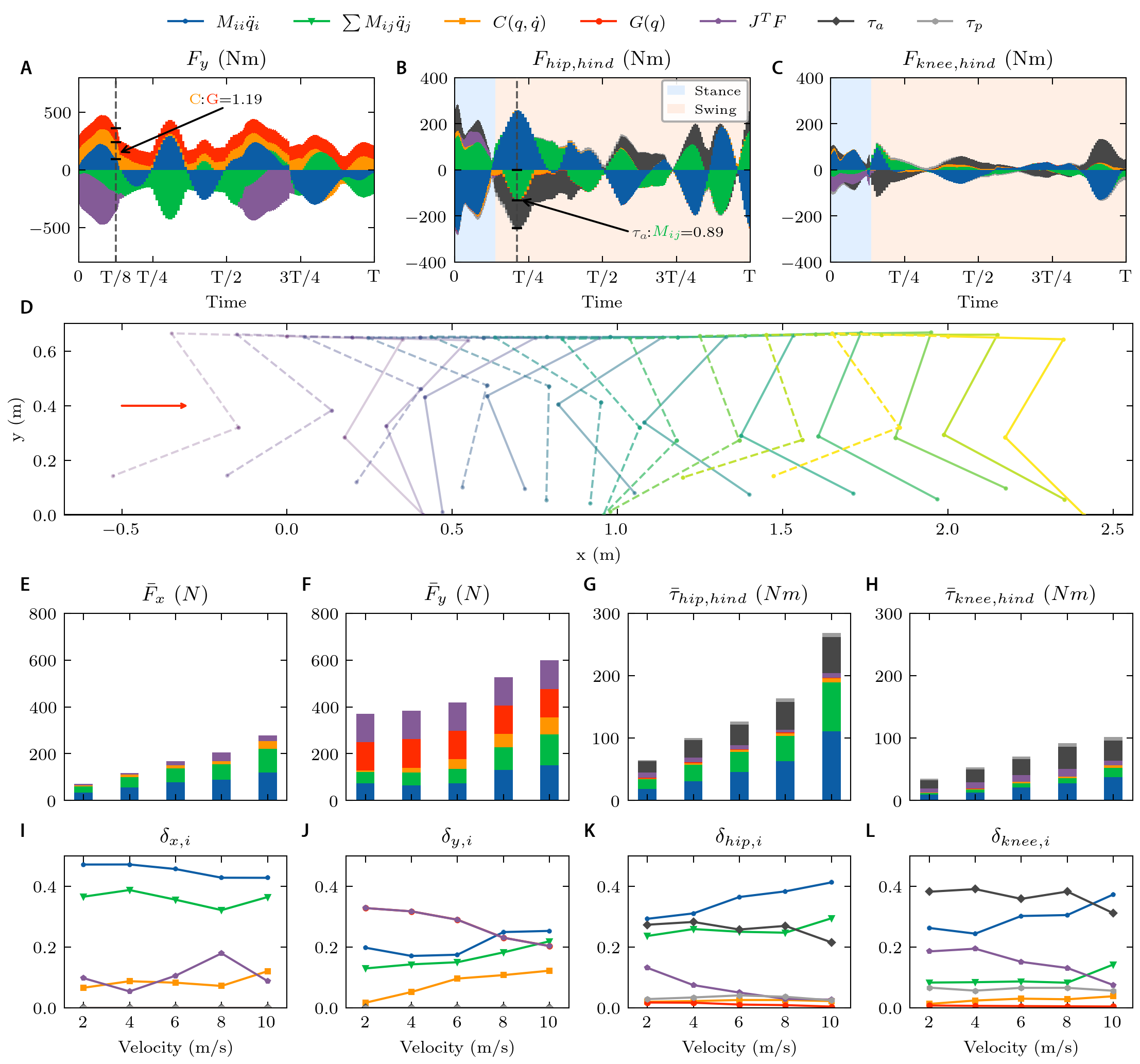}
	\caption{\textbf{Force decomposition and contribution analysis of the coupled dynamic model.}
	(\textbf{A-C}) Stacked force profiles at 10 m/s for key degrees of freedom. (\textbf{D}) Optimized robot trajectory at 10 m/s, with the red arrow indicating the direction of forward motion. (\textbf{E-H}) Mean magnitude of each force/torque component in different running speeds. (\textbf{I-L}) Relative contribution of each dynamic term in different running speeds.}
	\label{fig:robot_model_analysis}
\end{figure}

Based on fixed robot parameters (Table~\ref{tab:robot_params}), we generate parameterized periodic trajectories for different running speeds via Direct Collocation Method \cite{pardo_hybrid_2017}. To quantify the dynamic behavior during 10 m/s running, we analyze the force decomposition of the optimized gait using stacked force diagrams (Fig.~\ref{fig:robot_model_analysis}, A to C), which illustrate the instantaneous decomposition of all force components governing the robot’s force balance throughout the gait cycle.
Ground reaction forces are widely recognized as the primary determinant of high-speed running performance \cite{weyandFasterTopRunning2000}. However, the stacked force diagrams along the $y$-axis (Fig.~\ref{fig:robot_model_analysis}A) reveal that, at 10 m/s, the generation of these large contact forces cannot be attributed to gravity alone. Instead, multiple dynamic components contribute substantially. A representative case occurs at $t = 3/4T$, where $\mathbf{C}(\mathbf{q}, \dot{\mathbf{q}})$ alone, scaling directly with limb mass, can reach up to 1.19 times the gravitational load. This result highlights that the ability to generate large ground reaction forces at high speed is intrinsically linked to inertial effects.

Force analysis of hip joint (Fig.~\ref{fig:robot_model_analysis}B) illustrates that the
coupling effect effectively reduces the demand for motor torque during critical
phases of high-speed running. For example, at t = 0.21T, the thigh must rapidly
decelerate to switch from backward to forward swing, requiring substantial
torque to overcome main inertial torque $\mathbf{M}_{ii} \ddot{\mathbf{q}}_{i}$. The required
torque magnitude is approximately 250 N·m, which is 2.5 times the motor’s peak torque, implying that the actuator alone cannot generate sufficient torque.
At this moment, the coupled inertial force $\Sigma \mathbf{M}_{ij} \ddot{\mathbf{q}}_{j}$ aligns with the active hip torque $u$, thereby bridging the gap between the main inertial force and the active control force. This dynamic balancing occurs throughout the hip motion cycle, but its role differs between contact phases. During stance, the coupled inertial force provides the dominant contribution to balancing the main inertial-force demand, reducing the reliance on active actuation. During swing, the main inertial demand is conventionally expected to be supplied by active torque; however, near the demand peak, the coupled inertial force acts in the same direction as the active torque and supplies a substantial portion of the required balance. A similar pattern is observed at the knee joint (Fig.~\ref{fig:robot_model_analysis}C).

To quantify the contributions of different dynamic terms across running speeds, we define the average force $\bar{F}$ as the time-averaged magnitude of each force over one gait cycle.

\begin{equation}
	\bar{F}_i\equiv\frac{\int_0^T\left\|F_i\right\|dt}{T}
	\label{eq:average force} 
\end{equation}

\noindent
based on the calculated average force data (Fig.~\ref{fig:robot_model_analysis}, E to H),
we further define $\delta$ as the ratio of each dynamic component to the total average
force, representing its relative contribution:

\begin{equation}
	\delta_i\equiv\frac{\bar{F}_i T}{\sum\bar{F}_i T}=\frac{\bar{F}_i}{\sum\bar{F}_i}
	\label{eq:force ratio} 
\end{equation}

\noindent
Figs.~\ref{fig:robot_model_analysis}F and \ref{fig:robot_model_analysis}J show that, from 0 to 10 m/s, the mean magnitudes of gravitational force and the vertical ground reaction force remain identical, consistent with the conservation of vertical momentum over a gait cycle. In stark contrast, all motion-dependent inertial terms, including the main inertial force, coupled inertial force, centrifugal force, and Coriolis force, increase sharply with speed. This trend signifies a shift in the dominant factors of the system dynamics. As illustrated by the torque decomposition at the hip joint (Fig.~\ref{fig:robot_model_analysis}, G and K), while the absolute demand for active driving torque scales with speed, its relative contribution to the overall dynamics paradoxically declines. In contrast, the compensatory effect of coupled inertial forces becomes increasingly pronounced at higher velocities. This shift indicates that, in high-speed regimes, system performance is governed less by direct motor actuation and more by the coordinated interplay of inertial terms. Consequently, optimizing high-speed locomotion requires exploiting the full system dynamics rather than relying solely on augmenting hardware specifications.

\begin{figure}[!htbp]
	\centering
	\includegraphics[width=\textwidth,height=\textheight,keepaspectratio]{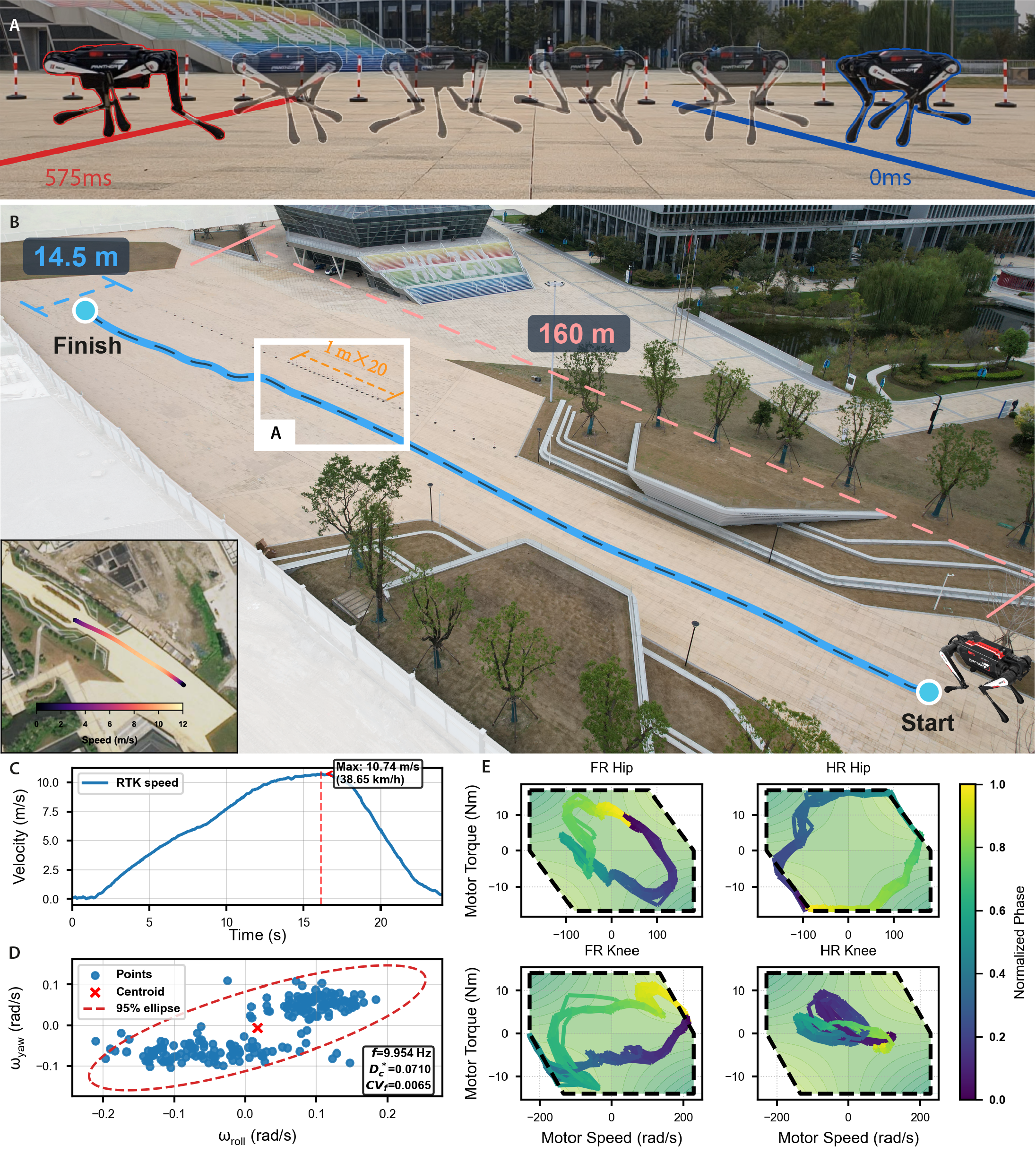}
	\caption{\textbf{Experimental results of high-speed locomotion.}
		(\textbf{A}) Representative keyframes of three consecutive gait cycles. Intermediate frames are faded and spatially shifted to illustrate the temporal evolution of leg configurations. (\textbf{B}) Bird's-eye view of the experimental environment. (\textbf{C}) GPS-recorded global velocity profile.  (\textbf{D}) Poincaré section of body angular-velocity dynamics during steady high-speed running. (\textbf{E}) Motor torque-speed operating points over three cycles during steady high-speed running.}
	\label{fig:Extreme_speed}
\end{figure}

\subsubsection*{Extreme-speed outdoor experiments}

Guided by our co-optimization framework, we design and manufacture the quadruped robot BP2. On conventional concrete pavement, BP2 achieves a maximum forward running speed of 10.74 m/s (Fig.~\ref{fig:Extreme_speed}C), setting a new outdoor speed record for quadrupedal robots (Fig.~\ref{fig:Extreme_speed}B). This performance represents a 22\% improvement over the previous record long held by WildCat \cite{WildCat}, despite BP2 having substantially smaller dimensions and relying on electric actuation.

Fig.~\ref{fig:Extreme_speed}A visualizes the steady-state running gait of BP2. The cycle begins with the touchdown of the right hind leg. Upon ground contact, the leg spring compresses to store a portion of the impact energy. Subsequently, the thigh initiates a push-off phase, and the spring reaches its maximum travel. As the leg completes its push-off phase, the robot enters the aerial phase—a critical period for high-speed locomotion in both robots and animals \cite{dai_whole-body_2014}. During this phase, the thigh rapidly reverses from backward to forward swing to achieve a sufficient stride length. The cycle concludes as the foot re-establishes contact with the ground, initiating the next stride (Movie~S2).

The stability of high-speed locomotion is evaluated via Poincaré section (Fig.~\ref{fig:Extreme_speed}D) analysis of trunk angular velocity. The return map remains tightly bounded, with minimal centroid drift ($D_c^* = 0.071$) and a high inlier ratio, indicating the absence of long-term divergence. Meanwhile, the motion is strongly phase-locked, with a dominant body frequency at twice the stride frequency driven by two ground impact events per gait cycle, and negligible variability ($CV_f = 0.006$). These results demonstrate stable, impact-driven limit-cycle behavior during high-speed running.

Onboard sensors record the torque-speed operating points of the thigh and knee motors for both the fore and hind limbs (Fig.~\ref{fig:Extreme_speed}E). The data show that even at extreme speeds, all motors operate consistently within their rated high-efficiency regions. Taking the hind limbs, which dominate power consumption, as an example, the peak power reaches 1620 W and 1765 W for the hind legs, respectively. These values are comparable to the reported peak motor power of the previous electrically actuated quadruped speed record holder HOUND (approximately 1706W), which achieves a top speed of 6.5 m/s \cite{shin_reinforcement_2025}. Using the dimensionless peak-performance metric defined in Fig.~\ref{fig:statistics}, BP2 reaches a ratio of 1.475, markedly higher than those of existing reported quadrupedal robots.  These results indicate that high-speed locomotion does not require disproportionate increases in motor power when system dynamics are effectively exploited. In particular, the hind-limb hip and front-knee joints fully utilize their speed-torque envelopes, validating the optimally selected gear reduction ratio, detailed in the Supplementary Materials (``Co-optimization of Hardware and Trajectory'' section).

\subsubsection*{Hundred-meter sprint benchmark}

Beyond peak velocity, sprint performance over a fixed distance provides a more comprehensive measure of overall capability, analogous to the 100-meter event in the Olympic Games. In this benchmark, BP2 is evaluated alongside a professional sprinter and untrained adults. The measured completion times are 11.2 s for the professional athlete, 12.2 s for BP2 (Movie~S3), and 16.1 s for untrained adults. These results show that BP2 substantially outperforms an untrained adult and approaches the performance level of a trained sprinter.

\begin{figure}[!htbp]
	\centering
	\includegraphics[width=\textwidth,height=0.58\textheight,keepaspectratio]{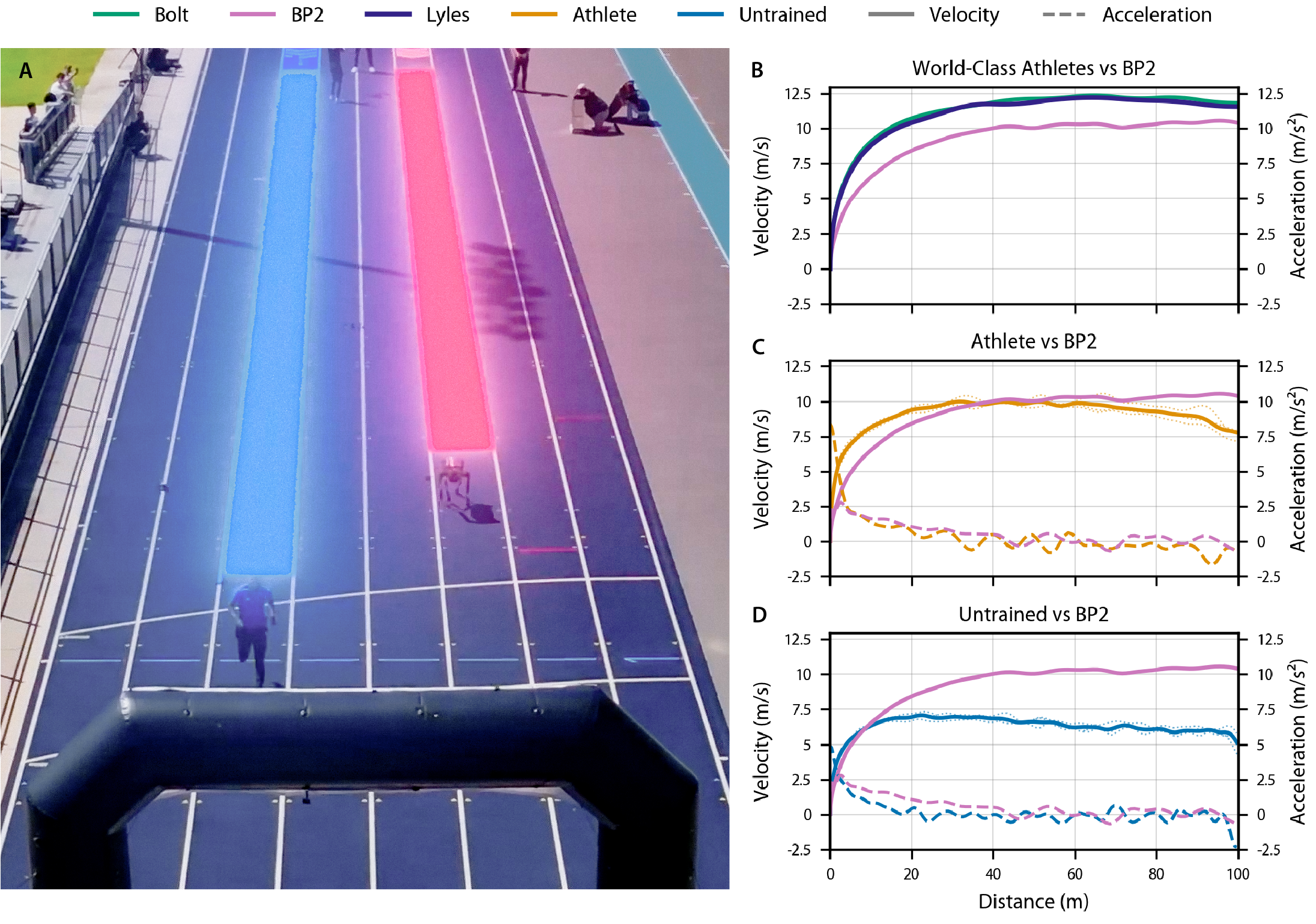}
	\caption{\textbf{Hundred-Meter Sprint Benchmark.}
		(\textbf{A}) Field snapshot of the race between BP2 and Lyles~\cite{mrbeastWorldsStrongestMan2025}.
		(\textbf{B}) Speed–distance profiles for BP2 and world-class athletes. The kinematic data for Noah Lyles and Usain Bolt are derived from the official 10 m split times reported for the 2024 Paris Olympics~\cite{OmegaAthleteFirst2024} and the biomechanical report of the 2009 Berlin World Championships~\cite{graubner2009biomechanical}, respectively, and the corresponding continuous curves are obtained through curve fitting.
		(\textbf{C}) Speed–distance profiles for BP2 and trained athletes. 
		(\textbf{D}) Speed–distance profiles for BP2 and untrained adult participants.}
	\label{fig:race}
\end{figure}

To further characterize sprint dynamics, we construct speed–distance profiles for all participants (Fig.~\ref{fig:race}, B to D). Across human subjects, a shared pattern emerges: an initial acceleration phase followed by a peak velocity that cannot be sustained to the finish line, with the onset and magnitude of speed decay varying with training level.
For untrained adults (Fig.~\ref{fig:race}D), the speed inflection point occurs between 20 and 40 meters, after which pronounced deceleration persists through the remainder of the race. The professional athlete (Fig.~\ref{fig:race}C) sustains acceleration over a longer distance, reaching a higher peak velocity before exhibiting a noticeable decline after approximately 60–70 meters, consistent with the onset of anaerobic fatigue and reduced neuromuscular recruitment efficiency. At the extreme end of human performance (Fig.~\ref{fig:race}B), the profiles of Usain Bolt (9.58 s, world record) and Noah Lyles (9.78 s) show that even elite sprinters undergo measurable deceleration over the final 20–30 meters, despite maintaining peak velocities in excess of 12 m/s. The performance gap between professional and elite sprinters in this phase is primarily attributed to differences in metabolic fatigue resistance and the ability to preserve neuromuscular coordination under acute physiological stress \cite{rossNeuralInfluencesSprint2001,Morin2015Hamstrings}.

In contrast, BP2 exhibits a qualitatively distinct speed profile (Fig.~\ref{fig:race}, B to D, red curves). After the acceleration phase, the robot maintains near-peak velocity with minimal decay over the remainder of the course, reflecting the sustained power output of its electric drive system, which is not limited by the metabolic constraints that govern biological sprinting. However, the dominant performance bottleneck lies in the initial 5-meter launch phase, where biological muscle provides superior explosive force generation. The trained sprinter and untrained adult achieve mean accelerations of 4.8 m/s\textsuperscript{2} and 2.8 m/s\textsuperscript{2}, respectively, whereas BP2 reaches 2.38 m/s\textsuperscript{2}, resulting in a time deficit of nearly 1 s relative to the professional athlete within the first 5 meters. This substantial acceleration deficit in the opening meters propagates into a time gap that persists throughout the race, despite BP2 achieving a peak speed comparable to or exceeding that of the professional athlete. These observations identify the initial acceleration phase, rather than peak speed or speed endurance, as the dominant performance-limiting factor for the current platform.

\begin{figure}[!htbp]
	\centering
	\includegraphics[width=\textwidth,height=0.58\textheight,keepaspectratio]{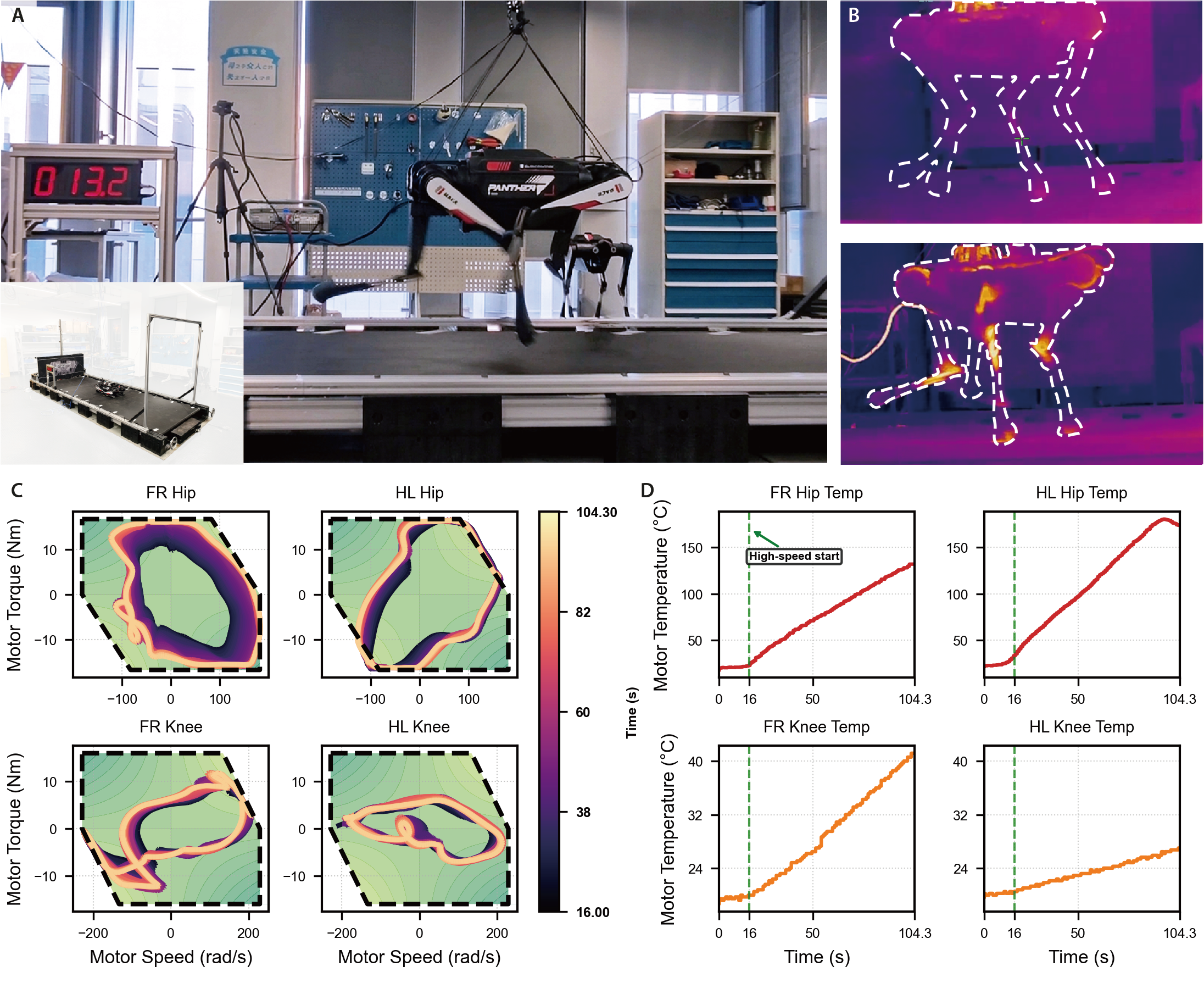}
	\caption{\textbf{Peak Speed Endurance Test.}
		(\textbf{A}) Experimental setup, snapshot of the robot at peak velocity. (\textbf{B}) Thermal images captured at the initial pose
		and during peak speed operation. (\textbf{C}) Torque-speed profile of the key
		joint motor during the endurance test. (\textbf{D}) Temperature curve of the key
		actuator during the endurance test.}
	\label{fig:treadmill}
\end{figure}

\subsubsection*{Peak speed sustained experiments}
To characterize the endurance limit underlying BP2's sustained high-speed locomotion, we conduct tests on a custom-built treadmill platform (Fig.~\ref{fig:treadmill}A). In these indoor treadmill experiments, BP2 reaches a peak running speed of 13.2 m/s (Movie~S4). In contrast to human sprinters, BP2 sustains running speeds above 10 m/s continuously for 86 s, confirming that the electric drive system can sustain near-peak velocity beyond short sprint intervals. However, this does not imply the absence of endurance limits; rather, the limiting mechanism shifts from biological fatigue to thermal-electromechanical constraints within the actuator system.

Actuator analysis shows that the primary motors operate close to their torque-speed saturation boundaries throughout the trial (Fig.~\ref{fig:treadmill}C). The time-resolved torque-speed trajectories, indicated by the color gradient, further reveal a progressive outward expansion of the motor operating cycles, indicating that the actuators progressively approach the boundary of their feasible operating envelope. This evolution results in a continuously decreasing torque margin for disturbance rejection and gait regulation. Thermal measurements show that heat accumulation is concentrated in the motors and series-elastic transmission elements (Fig.~\ref{fig:treadmill}B). In the most heavily loaded rear hip joint, temperature increases nearly linearly at approximately 1.8 °C/s during the 86 s trial. This thermal accumulation weakens motor magnetic flux linkage, reduces the effective torque constant, and increases winding resistance, thereby decreasing available torque while the gait continues to demand operation near actuator limits. Once the torque margin is exhausted, the robot can no longer maintain the periodic ground reaction forces required for high-speed balance. These results show that sustained high-speed quadrupedal locomotion is ultimately bounded by thermal regulation and torque-margin preservation, rather than by peak-speed generation alone.

\subsection*{DISCUSSION}
\noindent
This study shows that high-speed quadrupedal running can be advanced by understanding and exploiting the internal coupled dynamics of the robot body. Motivated by Huygens' coupled pendulums, we identify inter-limb inertial coupling as a mechanism that redistributes energy during rapid periodic motion and reduces the peak joint torque required to sustain high-speed locomotion. Trajectory optimization is used as an analytical tool to examine this mechanism in quadrupedal running; by further introducing hardware parameters as optimization variables, the analysis is extended from motion generation to hardware design. This framework leads to the development of BP2, which reaches 10.74 m/s in outdoor experiments, representing the fastest recorded speed for a quadrupedal robot, and provides a pathway from physical insight to high-performance robot design.

A central implication of these results is that limb inertia should not be viewed solely as a parasitic load in high-speed legged locomotion. Although reducing limb inertia can simplify control and decrease swing effort, the present results show that appropriately distributed inertia can actively participate in whole-body energy exchange and reduce instantaneous actuation demands. In this sense, high-speed performance depends not only on actuator capacity or controller bandwidth, but also on how the mechanical body shapes the flow of energy and momentum across the gait cycle. The agreement between optimized motion patterns, actuator-level measurements, and real-world running performance supports inertial coupling as a physically meaningful mechanism rather than an incidental outcome of numerical optimization.

Several limitations remain. First, the dynamic model and theoretical analysis in this study are primarily formulated in the sagittal plane. This assumption is appropriate for straight-line sprinting, but it does not capture the full three-dimensional dynamics required for turning, lateral maneuvers, or rapid transitions in unstructured environments. Importantly, the optimization strategy itself is not restricted to two-dimensional motion: the same formulation can be extended by incorporating lateral dynamics, yaw motion, roll-pitch coupling, and three-dimensional contact constraints. Such an extension would allow the inertial coupling mechanism identified here to be examined in a broader class of agile locomotor behaviors beyond straight-line running. Second, the present work focuses on maximum running speed as the primary performance metric. The same mechanism-guided analysis could be extended to other objectives, including jumping, payload transport, endurance, and energy efficiency. Third, the optimization analysis and controller training are currently performed as separate stages. Future work could use the optimized trajectories as physical priors for learning-based control and improving exploration efficiency and performance in high-speed regimes.

\subsection*{MATERIALS AND METHODS}
\subsubsection*{Overview}
The study follows four main steps. We first use a Huygens' coupled pendulum model to examine inertial coupling in a simplified setting. We then formulate a hybrid dynamics model for sagittal-plane quadrupedal running with intermittent ground contact. Based on this model, trajectory optimization and hardware-parameter co-optimization are used to identify feasible high-speed gaits and guide the design of BP2. The resulting robot is tested in outdoor peak-speed and 100-meter sprint experiments, as well as treadmill-based sustained high-speed running trials.

\subsubsection*{Huygens' coupled pendulum model}
\noindent
To isolate the basic mechanism of inertial coupling, we first consider a simplified two-pendulum system inspired by Huygens' coupled clocks. The model consists of two pendulums attached to a shared translating beam. Each pendulum has mass $m$ and length $l$, and the beam has mass $M$. The horizontal displacement of the beam is denoted by $x$, and the angular positions of the two pendulums are denoted by $\theta_1$ and $\theta_2$.

This reduced model is not intended to represent the full quadrupedal robot. Instead, it provides a minimal dynamical system in which coupling-mediated energy exchange can be analyzed explicitly.

Under these assumptions, the equations of motion are written as follows:
\begin{equation}
\begin{bmatrix}
    M+2m & m l C_{1} & m l C_{2} \\
    m l C_{1} & m l^{2} & 0 \\
    m l C_{2} & 0 & m l^{2}
\end{bmatrix}
\begin{bmatrix}
    \ddot x \\
    \ddot{\theta}_{1} \\
    \ddot{\theta}_{2}
\end{bmatrix}
+
\begin{bmatrix}
    -m l \dot{\theta}_{1}^{2} S_{1}-m l \dot{\theta}_{2}^{2} S_{2} \\
    0 \\
    0
\end{bmatrix}
+
\begin{bmatrix}
    0 \\
    m g l S_{1} \\
    m g l S_{2}
\end{bmatrix}
=
\begin{bmatrix}
    0 \\
    \tau_{1} \\
    \tau_{2}
\end{bmatrix}
\label{eq:coupled_dynamics}
\end{equation}

\noindent
to simplify the analysis, we introduce the following dimensionless parameters. We define the dimensionless displacement as $\sigma = x/l$ and dimensionless time as $\omega_0 t$, where $\omega_0=\sqrt{g/l}$ is the natural frequency of the system. The dimensionless output torque is defined as $u_i = \tau_i/mgl$, and the mass ratio parameter as $R = M/m$. Substituting these variables into the original dynamics yields the following dimensionless form:
\begin{equation}
\begin{bmatrix}
    R + 2 & C_{1} & C_{2} \\
    C_1 & 1 & 0 \\
    C_2 & 0 & 1
\end{bmatrix}
\begin{bmatrix}
    \sigma'' \\
    {\theta}''_1 \\
    {\theta}''_2
\end{bmatrix}
+
\begin{bmatrix}
    - \theta_1'^2 S_1 - \theta_2'^2 S_2 \\
    0 \\
    0
\end{bmatrix}
+
\begin{bmatrix}
    0 \\
    S_{1} \\
    S_{2}
\end{bmatrix}
=
\begin{bmatrix}
    0 \\
    u_{1} \\
    u_{2}
\end{bmatrix}
\label{eq:dimensionless}
\end{equation}

\noindent
where the prime symbol ( $'$ ) denotes differentiation with respect to dimensionless time. 

If the beam translation is constrained, i.e., $\sigma=\sigma'=\sigma''=0$, the system becomes dynamically equivalent to two independent pendulums, and the corresponding torque requirement is defined as $\tilde{u}_i=\theta_i''+S_i$. When the beam is free to translate, however, the required actuation torque for each pendulum depends not only on its own motion, but also on the motion of the other pendulum through the shared beam. This coupling gives rise to the additional torque term in Eq.~\ref{eq:active torque}, which captures the redistribution of inertial effects between the two oscillators.

\subsubsection*{Sagittal Plane hybrid dynamic model}
\noindent
To analyze straight-line high-speed running while retaining the dominant body-limb coupling effects, we formulate the robot dynamics in the sagittal plane (Fig.~\ref{fig:model}). The trunk is modeled as a finite-length rigid body with horizontal and vertical translation and pitch rotation; fore and hind hip positions are defined relative to the pitching trunk, so that the model preserves the influence of body geometry and trunk rotation on limb motion and ground contact. Assuming a left-right symmetric trotting gait, the two diagonal leg pairs are represented by equivalent sagittal-plane fore and hind limbs with a half-cycle phase shift. This formulation captures the continuous running dynamics and intermittent contact events relevant to high-speed straight-line locomotion.

The generalized coordinates of legged robots can vary across contact states if stance and swing dynamics are modeled with phase-specific coordinate sets \cite{sutrisno_how_2020}. Although such formulations reduce the number of degrees of freedom in each phase, they require coordinate transformations at contact transitions and lead to different equation forms across phases. To maintain a consistent representation, we use a unified set of generalized coordinates and impose contact conditions through geometric constraints and contact forces \cite{xi_selecting_2016,featherstone_rigid_2008,hwangbo_per-contact_2018}. This formulation corresponds to constrained rigid-body dynamics with Lagrange multipliers and provides a standardized equation form for trajectory optimization and simulation.

At touchdown, the system undergoes an instantaneous velocity transition caused by the contact impulse. Integrating the continuous equations of motion over the infinitesimal impact interval $[t_c,t_c+\Delta t]$ gives:
\begin{equation}
\int_{{t_c}}^{{t_c} + \Delta t} {({\bf{M}}({\bf{q}}){\bf{\ddot q}} 
+ {\bf{C}}({\bf{q}},{\bf{\dot q}}) + {\bf{G}}({\bf{q}})g)dt 
= \int_{{t_c}}^{{t_c} + \Delta t} ({\bf{\tau }}_a} + {{\bf{\tau }}_p} + {{\bf{J}}^T}{\bf{F}}) dt
\end{equation}

\noindent
as $\Delta t \rightarrow 0$, the finite Coriolis, gravitational, and joint-torque contributions vanish over the impact interval, whereas the contact impulse remains finite. Assuming the configuration does not change during impact, the impact map becomes:

\begin{equation}
\begin{array}{c @{\quad} c @{\quad} c}
{\bf{M}}({\bf{q}})\int_{{t_c}}^{{t_c} + \Delta t} {{\bf{\ddot q}}dt = {{\bf{J}}^T}\int_{{t_c}}^{{t_c} + \Delta t} {\bf{F}} } dt
& \Rightarrow &
{\bf{M}}({\bf{q}})\Delta {\bf{\dot q}} = {{\bf{J}}^T}\Lambda 
\end{array}
\end{equation}
\noindent
where $\Delta{\bf{\dot q}}$ is the jump in generalized velocity and $\boldsymbol{\Lambda}=\int_{t_c}^{t_c+\Delta t}{\bf{F}}dt$ denotes the contact impulse. Combining this impact map with the continuous constrained dynamics yields the complete hybrid dynamic model,

\begin{equation}
\left\{\begin{array}{ll}
\mathbf{M}(\mathbf{q}) \ddot{\mathbf{q}}+\mathbf{C}(\mathbf{q}, \dot{\mathbf{q}})+\mathbf{G}(\mathbf{q}) g=\boldsymbol{\tau}_{a}+\boldsymbol{\tau}_{p}+\mathbf{J}^{T} \mathbf{F}, 
& t \notin \{t_{c0},t_{c1},\cdots\}, \\[4pt]
\left\{\begin{array}{l}
\mathbf{q}^{+}=\mathbf{q}^{-}, \\
\mathbf{M}(\mathbf{q})\left(\dot{\mathbf{q}}^{+}-\dot{\mathbf{q}}^{-}\right)=\mathbf{J}^{T}\boldsymbol{\Lambda},
\end{array}\right.
& t \in \{t_{c0},t_{c1},\cdots\}.
\end{array}\right.
\label{eq:hybrid}
\end{equation}

\noindent
here, superscripts $-$ and $+$ indicate quantities immediately before and after impact, respectively. The continuous part of this model is used in the Results section to analyze inertial coupling and actuation demand during periodic running, whereas the full hybrid formulation, including the impact relation above, is used for trajectory optimization and simulation.

\subsubsection*{Trajectory Optimization and Hardware Optimization}
\noindent
To analyze the dynamics of high-speed quadrupedal running and translate the resulting insights into hardware design, we formulate the optimization problem using the Direct Collocation Method \cite{xi_optimal_2014,pardo_hybrid_2017}. The hybrid dynamics model are discretized over a periodic gait cycle, converting the trajectory generation problem into a large-scale nonlinear programming (NLP) problem. The same framework serves two complementary purposes. First, with hardware parameters fixed, trajectory optimization serves as an analytical tool for examining how inertial coupling affects joint torque demand during rapid periodic locomotion. Second, selected hardware parameters are introduced as additional optimization variables, extending the analysis from motion generation to high-performance robot design.

To obtain a differentiable formulation suitable for gradient-based optimization, foot-ground contact is represented using continuous contact variables and constraints. The gait cycle is divided into four segments with a fixed number of collocation points in each segment, whereas the duration of each segment is treated as an optimization variable. This formulation allows the optimizer to determine gait cycle period and duty factor without explicitly enumerating discrete hybrid events.

The optimization variables for the trajectory problem include the generalized coordinates $\mathbf{q}_i^t$, generalized velocities $\dot{\mathbf{q}}_i^t$, active joint torques $\boldsymbol{\tau}_{\mathrm{a},i}^t$, ground reaction forces $\mathbf{F}_i^t$, and the segment durations $(T_1,T_2,T_3,T_4)$. Here, $i$ indexes the four gait segments, and $t$ indexes the discrete nodes within each segment. Each segment duration $T_i$ is divided into a fixed number of $N$ discretization intervals, with the time step defined as $\delta t_i=T_i/N$. The segment durations $T_i$ are optimized, whereas $N$ is fixed.  The hardware-optimized formulation uses the same trajectory-level cost structure while additionally allowing selected mechanical parameters to vary. For clarity, the objective function of the initial fixed-hardware trajectory optimization stage is given below; it is formulated as a weighted sum of normalized node-wise costs penalizing mechanical power, actuator and contact loading, and temporal nonsmoothness:

\begin{equation}
\begin{aligned}
\min_{\mathbf{q}_i^t,\dot{\mathbf{q}}_i^t,
\boldsymbol{\tau}_{a,i}^t,\mathbf{F}_i^t,T_i}
\quad
J
&=
\sum_{i=1}^{4}\sum_{t=0}^{N-1} R_i^t .
\end{aligned}
\label{eq:obj_func}
\end{equation}

\noindent
The running cost $R_i^t$ is defined as

\begin{equation}
\begin{array}{r@{\;}c@{\;}l}
R_i^t
& = &
\omega_p
\underbrace{
\sum\limits_{\ell=1}^{4}
\left(
\dfrac{
\dot q_{\ell,i}^t \tau_{\mathrm{a},\ell,i}^t
}{
\dot q^{\max}\tau^{\max}
}
\right)^2
}_{R_{p,i}^t}
\\[6pt]
& + &
\omega_f
\underbrace{
\left[
\sum\limits_{\ell=1}^{4}
\left(
\dfrac{
\tau_{\mathrm{a},\ell,i}^t
}{
r_\ell \tau^{\max}
}
\right)^2
+
\sum\limits_{j=1}^{4}
\left(
\dfrac{
F_{j,i}^t
}{
\bar{F}
}
\right)^2
\right]
}_{R_{\tau F,i}^t}
\\[6pt]
& + &
\omega_s
\underbrace{
\left[
\sum\limits_{\ell=1}^{4}
\left(
\dfrac{
\tau_{\mathrm{a},\ell,i}^t-\tau_{\mathrm{a},\ell,i}^{t-1}
}{
r_\ell \tau^{\max}
}
\right)^2
+
\sum\limits_{j=1}^{4}
\left(
\dfrac{
F_{j,i}^t-F_{j,i}^{t-1}
}{
\bar{F}
}
\right)^2
\right]
}_{R_{s,i}^t}.
\end{array}
\label{eq:R}
\end{equation}
\noindent
where $\ell$ and $j$ index the actuated joints and the contact-force components, respectively. The variable $r_\ell$ is the corresponding gear ratio. The quantities $\dot q^{\max}$ and $\tau^{\max}$ denote the motor-side maximum speed and maximum torque, respectively, while $\bar F$ is the reference contact force used for normalization. The three components of the objective penalize mechanical power ($R_p^t$), actuator and contact loading ($R_\tau^t$), and motion nonsmoothness ($R_s^t$). The optimization does not explicitly impose inertial coupling as an objective; rather, coupling effects emerge from dynamically feasible high-speed solutions with reduced actuator loading.

The NLP is subject to five categories of constraints: physical limits, kinematic constraints, gait periodicity, task-specific requirements, and actuator saturation. These constraints ensure that the optimized trajectories satisfy the robot dynamics, contact feasibility, prescribed running speed, and hardware limits. Detailed mathematical formulations of the constraints are provided in the Supplementary Materials (section S1.2).

All optimization problems are implemented and solved using \textsc{CasADi} \cite{andersson_casadi_2019}. The fixed-hardware trajectory optimization is used to analyze the relationship between inertial coupling and joint torque demand. The hardware-parameter-augmented optimization then provides candidate mechanical parameters and corresponding running trajectories for guiding the design of BP2.

\subsubsection*{Experimental Platform and Setup}
\noindent

We develop and manufacture the quadruped robot BP2 as the experimental platform. As shown in Fig.~\ref{fig:bp2}, BP2 is designed with several structural features for high-speed running. The center of mass is lowered to improve heading regulation and reduce yaw deviation during rapid locomotion. The forelimb track width is increased relative to the hindlimb track width, reducing the risk of self-collision between the front and rear legs under the large stride lengths required at high speed. Each leg incorporates two parallel springs with a stiffness of 3000 N/m each, providing elastic energy storage during impact and reducing the power burden on the actuators. The foot design is inspired by cheetah paws and sprinting shoes, with cleats distributed over a contact envelope exceeding 60 degrees to maintain traction across varying touchdown angles. Each joint is actuated by a frameless DC motor paired with a low-ratio planetary gearbox, providing high torque output while preserving backdrivability. To satisfy stiffness requirements while maintaining the optimized mass distribution, the main frame and leg linkages are fabricated from lightweight aluminum alloy and Ti-6Al-4V titanium alloy. Proprioceptive sensing is provided by absolute joint encoders and an Xsens MTi-630 nine-axis inertial measurement unit. Onboard computation is performed on an UP Board with an Intel Atom x5-Z8350 processor. The main controller operates at 200 Hz, and the low-level force controller is updated at 1 kHz. Joint encoders, the inertial measurement unit, and motor current sensors record onboard data at 1 kHz.

For the real-world experiments, BP2 is controlled using the reference-motion-guided and relaxed reinforcement learning policy \cite{jin_high-speed_2022}, with reward terms extended for the higher-speed regime considered here. Details of the training procedure are provided in the ``RL training'' section of the Supplementary Materials.

High-speed running experiments are conducted in outdoor environments, including a synthetic track and concrete surface. Ground flatness for record-speed trials is verified using laser measurement. Global velocity is measured using an external high-speed tracking system and cross-validated with an onboard real-time kinematic GNSS module, both sampled at 10 Hz. Indoor sustained high-speed running experiments are conducted on a custom-built treadmill platform, with belt speed measured by a roller mechanically coupled to the treadmill belt.



\clearpage 

%
\bibliography{bp2speed}
\bibliographystyle{sciencemag}

%
%
%
%
%
%


\section*{Acknowledgments}
\paragraph*{Funding:}
This work was supported by grants from the STI2030-Major Projects No. 2025ZD0218800
\paragraph*{Author contributions:}
Y.T. and Y.J. conceived the main research idea, conducted the experiments, and contributed to the overall system development. Y.T. was responsible for trajectory optimization, hardware parameter optimization and analysis, controller training and evaluation, and manuscript writing. Y.J. led the iterative hardware design of the robot and revised the manuscript. S.C. assisted with controller training and experimental testing. X.L. and G.L. contributed to the construction of the testing platform and assisted with hardware iteration. Y.Y. assisted with experimental testing, trajectory optimization, and hardware parameter optimization analysis. C.S. assisted with trajectory optimization and hardware parameter optimization analysis. Y.L. assisted with experimental testing, photography, and video editing. C.F. assisted with software debugging. W.Y. and H.W. initiated the project, provided financial support, and revised the manuscript.
\paragraph*{Competing interests:}
The authors declare that they have no competing interests.
\paragraph*{Data and materials availability:}
All data needed to evaluate the conclusions in the paper are present in the main text or the Supplementary Materials. The RL controller code are available at \url{https://github.com/WoodenJin/High_Speed_Quadrupedal_Locomotion_by_IRRL}.


\subsection*{Supplementary materials}
Materials and Methods\\
Result\\
Figs. S1 and S2\\
Tables S1 to S4\\
References \textit{(52 - \arabic{enumiv})}\\

\subsubsection*{Other Supplementary Materials for this manuscript:}
Movies S1 and S4\\


\newpage


\renewcommand{\thefigure}{S\arabic{figure}}
\renewcommand{\thetable}{S\arabic{table}}
\renewcommand{\theequation}{S\arabic{equation}}
\renewcommand{\thepage}{S\arabic{page}}
\setcounter{figure}{0}
\setcounter{table}{0}
\setcounter{equation}{0}
\setcounter{page}{1} 

\begin{center}
\section*{Supplementary Materials for\\ \scititle}

Yucheng Tao et al.\\[2em]
\small Corresponding author: Yongbin Jin, yongbinjin@zju.edu.cn\\
\end{center}

\subsubsection*{This PDF file includes:}
Materials and Methods\\
Result\\
Figs. S1 and S2\\
Tables S1 to S4\\
References \textit{(52-\arabic{enumiv})}\\

\subsubsection*{Other Supplementary Materials for this manuscript:}
Movies S1 and S4\\

\newpage

\subsection*{Materials and Methods}

\subsubsection*{Constraints of co-optimization framework}
\noindent
Utilizing a finite difference formulation, we discretize Equation~\ref{eq:hybrid} in the time domain to derive the physical constraints, as follows:

\begin{equation}
\begin{array}{*{20}{l}}
{\left\{ {\begin{array}{*{20}{l}}
{{{\bf{q}}^{t + 1}} - {{\bf{q}}^t} - \frac{{\Delta t}}{2}\left( {{{{\bf{\dot q}}}^{t + 1}} + {{{\bf{\dot q}}}^t}} \right) = {\bf{0}}}\\
{{\bf{M}}({\bf{q}}^t)\frac{{{{{\bf{\dot q}}}^{t + 1}} - {{{\bf{\dot q}}}^t}}}{{\Delta t}} + {\bf{C}}\left( {{{\bf{q}}^t},{{{\bf{\dot q}}}^t}} \right) + {\bf{G}}\left( {{{\bf{q}}^t}} \right) - {\bf{B}}\boldsymbol{\tau} _a^t - \boldsymbol{\tau} _p^t - {{\bf{J}}^T}\left( {{{\bf{q}}^t}} \right){{\bf{F}}^t} = {\bf{0}}}
\end{array}} \right.}&{\forall t \notin \{t_{c0},t_{c1},\cdots\}}\\
{\left\{ {\begin{array}{*{20}{l}}
{{{\bf{q}}^{t,+ }} - {{\bf{q}}^{t,- }} = {\bf{0}}}\\
{{\bf{M}}\left( {{{\bf{q}}^{t, - }}} \right)\left( {{{{\bf{\dot q}}}^{t, + }} - {{{\bf{\dot q}}}^{t, - }}} \right) - {{\bf{J}}^T}\left( {{{\bf{q}}^{t, - }}} \right){\boldsymbol{\Lambda}^{t}} = {\bf{0}}}
\end{array}} \right.}&{\forall t \in \{t_{c0},t_{c1},\cdots\}}
\end{array}
\label{eq:Physical Constraints}
\end{equation}
\noindent
where ${\bf B}$ is the actuator selection matrix that maps the active joint torques $\boldsymbol{\tau}_a^t$ into the generalized force space.

Our optimization framework is constructed within a symbolic architecture.
Therefore, this work necessitates a predefined contact sequence between the
robot's feet and the ground over a full cycle. Based on this contact sequence,
the legged locomotion constraints are derived as follows:

\begin{equation}
\begin{array}{ll}
\left\{
\begin{array}{l}
\phi_y(\mathbf{q}^{t})=0,\\
\phi_x(\mathbf{q}^{t})=\phi_x(\mathbf{q}^{0}),\\
F_y^{t}\ge 0,\\
|F_x^{t}|\le \mu F_y^{t}
\end{array}
\right.
&
\text{Stance phase},
\\[18pt]
\left\{
\begin{array}{l}
\phi_y(\mathbf{q}^{t})\ge 0,\\
F_x^{t}=F_y^{t}=0
\end{array}
\right.
&
\text{Swing phase}.
\end{array}
\label{eq:Motion Constraints}
\end{equation}

\noindent
where $\phi_x$ and $\phi_y$ denote the horizontal and vertical positions of the corresponding foot in the world frame. The same stance and swing constraints are applied to each foot according to the prescribed contact sequence. To prevent relative slip between the foot and the ground, the tangential and normal contact forces are constrained by the friction cone with coefficient $\mu$.

In addition to the constraints mentioned above, we also impose periodic
constraints, task-specific constraints, boundary constraints and smoothness constraints.

\begin{equation}
\left\{ {\begin{array}{*{20}{l}}
{\left\{ {\begin{array}{*{20}{l}}
{{q^0} - {q^N} = 0,\quad (\mathrm{except} \quad x)}\\
{{{\dot q}^0} - {{\dot q}^N} = 0}
\end{array}} \right.}&{{\rm{Periodic~Constraint}}}\\
{}&{}\\
{\left\{ {\begin{array}{*{20}{l}}
{{q_x^0}=0}\\
{{q_x^N} - {q_x^0} - V_{{\rm{target}}}\sum_{i=1}^{4}T_i = 0}
\end{array}} \right.}&{{\rm{Task~Constraint}}}\\
{}&{}\\
{\left\{ {\begin{array}{*{20}{l}}
{ \underline{q} \le {q^t} \le \bar q}\\
{ \underline{{\dot q}} \le {{\dot q}^t} \le \bar {\dot q}}\\
{\underline{\tau}(\dot q^t) \le {\tau_a ^t} \le \bar \tau (\dot q^t)}
\end{array}} \right.}&{{\rm{Boundary~Constraint}}}\\
{}&{}\\
{\left\{ {\begin{array}{*{20}{l}}
{\left| \tau_a^{t+1}-\tau_a^t \right| \le \Delta \tau_{\max}}\\
{\left| {F}^{t+1}-{F}^{t} \right| \le \Delta F_{\max}}
\end{array}} \right.}&{{\rm{Smoothness~Constraint}}}
\end{array}} \right.
\label{other constraints}
\end{equation}

\noindent
where $\bar{q},\underline{q},\bar{\dot q},\underline{\dot
q},\bar{\tau},\underline{\tau}$ denote the upper and lower bounds on the joint
generalized coordinates and velocities, and torques, respectively.
The upper and lower limits of the motors are determined by the torque-speed
curve of the back electromotive force at high speeds.

\subsubsection*{RL training}
\noindent
A kinematic gait generator provides reference joint trajectories and joint velocities for commanded forward, lateral, and yaw velocities. During the initial training stage, the policy is guided by reference-motion imitation terms to obtain dynamically feasible periodic gaits. The imitation constraints are then relaxed by increasing the relative importance of velocity tracking and regularization terms, allowing the policy to deviate from the reference trajectory when such deviations improve high-speed stability and actuator feasibility.

The policy is trained with proximal policy optimization (PPO) in the RaiSim simulation environment \cite{hwangbo_per-contact_2018}. The neural network receives the commanded velocity, gait phase, joint states, body orientation, and body angular velocity as observations, and outputs joint-level position targets for a low-level PD controller. The same training formulation is used for all controllers in this study. To adapt the method to the higher-speed regime of BP2, the reward design is extended in two ways. First, the relative weighting between motion imitation and command tracking is made speed-dependent, reducing the constraint imposed by the reference trajectory at high commanded velocities. Second, an additional straight-running symmetry reward is introduced to suppress unnecessary abduction/adduction motion when the commanded lateral and yaw velocities are small. The complete reward terms are summarized in Table~\ref{tab:rl_reward}.

\subsection*{Result}

\subsubsection*{Co-optimization of Hardware and Trajectory}

\begin{figure}
	\centering
	\includegraphics[width=1.0\textwidth]{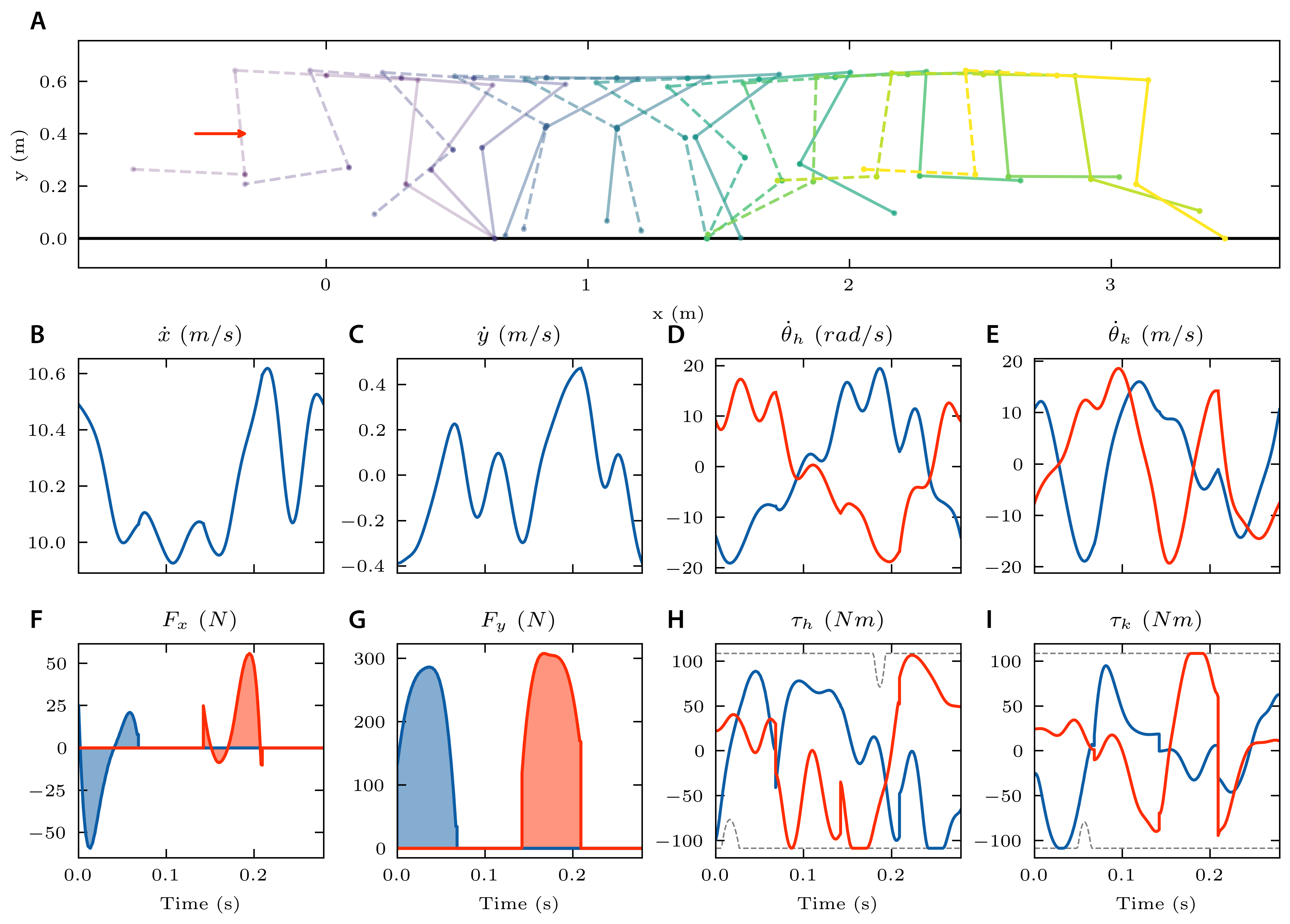}
	\caption{\textbf{Co-optimization results for the sagittal-plane model at 10 m/s.}
		Throughout, blue and red curves represent the left and right legs, respectively.
		(\textbf{A}) Robot trajectory, with the red arrow indicating forward motion. (\textbf{B-E}) Velocity profiles for horizontal center-of-mass (CoM) motion (B), vertical CoM motion (C), hip rotation (D), and knee rotation (E). (\textbf{F, G}) Ground reaction forces. (\textbf{H, I}) Hip (H) and knee (I) joint torques. Gray dashed lines indicate the upper and lower torque limits of the hind leg motor at the corresponding joint velocities.}
	\label{fig:hardware_opt}
\end{figure}

Fig.~\ref{fig:hardware_opt} shows the resulting optimized motion for the sagittal-plane model at a target speed of 10 m/s. The trajectory over one gait cycle exhibits a periodic running pattern with alternating contact phases (Fig.~\ref{fig:hardware_opt}A). The corresponding velocity profiles show that the optimized motion remains continuous across the cycle, with the elastic elements reducing the abrupt velocity changes associated with foot-ground impact (Fig.~\ref{fig:hardware_opt}, B to E).

The optimized ground reaction forces satisfy the impulse requirements for periodic locomotion (Fig.~\ref{fig:hardware_opt}, F and G). Over one complete gait cycle, the net horizontal impulse approaches zero, preserving periodic horizontal motion, whereas the net vertical impulse balances the gravitational impulse required to maintain periodic vertical motion. The active hip and knee torques remain within the torque-speed limits of the selected motors (Fig.~\ref{fig:hardware_opt}, H and I), indicating that the optimized trajectory is dynamically feasible under the prescribed actuator constraints.

With fixed motor specifications, the formulation can incorporate multiple hardware parameters as optimization variables. In this example, we select three key parameters that directly affect actuation capacity, morphology, and elastic energy storage: gear reduction ratio, leg length, and spring stiffness. The optimization yields a gear reduction ratio of 8.5, a leg length of 0.35 m, and a spring stiffness of 6000 N/m. These values provide quantitative guidance for the hardware design by linking the optimized high-speed gait to implementable mechanical parameters.


\begin{figure}[h] 
	\centering
	\includegraphics[width=1.0\textwidth]{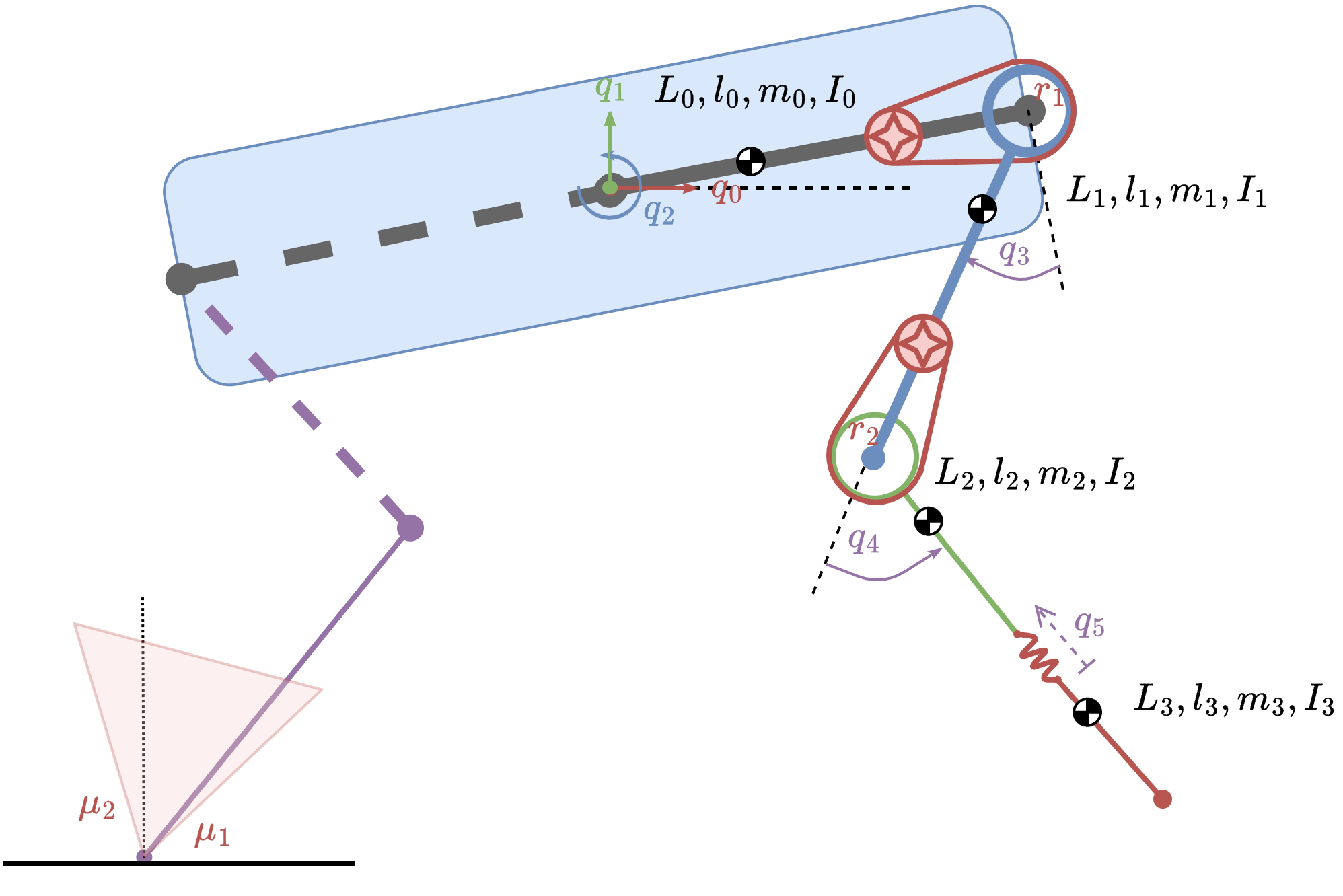} 

	\caption{\textbf{Hybrid dynamic model of the spring-legged robot with a complete motor inertia model.}
	Schematic of the sagittal-plane spring-legged model used for deriving the hybrid dynamics. The model is used for the forelimbs and adapted for the hindlimbs by setting $L_0$ and $l_0$ to negative values. In the derived equations of motion, the motor- and transmission-related parameters enter only through the rotor inertia $I_m$ and the gear ratio $r$.}
	\label{fig:model} 
\end{figure}

\clearpage

\begin{table}[h] 
	\centering
	\caption{\textbf{Performance specifications of representative legged robots.}
		The table summarizes the body mass, motor peak torque, maximum motor speed, and achievable peak locomotion speed of different robotic platforms.}
	\label{tab:robot_specs}

	\begin{tabular}{ccccc}
		\hline
		Robot Model & Body Mass & Motor Peak Torque & Motor Max Speed & Peak Speed \\
					& $m$ (kg)  & $\tau^{max}$ (Nm)     & $\omega_{max}$ (rad/s) & $v$ (m/s) \\
		\hline
		MIT Mini Cheetah\cite{katz_mini_2019} & 9 & 18 & 40 & 3.9\\
		BlackPanther\cite{jin_high-speed_2022} & 10 & 18 & 42 & 5\\
		\textbf{BlackPanther2} & \textbf{36} & \textbf{120} & \textbf{21.4} & \textbf{10.74}\\
		StarlETH\cite{hutterSTARLETHCOMPLIANTQUADRUPEDAL2012a} & 23 & 24 & 10.5 & 0.75\\
		ANYmal\cite{hwangboLearningAgileDynamic2019} & 30 & 40 & 12 & 1.6\\
		MIT Cheetah II\cite{park_high-speed_2017} & 34.4 & 174 & 21 & 6.4\\
		MIT Cheetah III\cite{bledt_mit_2018} & 45 & 230 & 21 & 3\\
		WildCat\cite{WildCat} & 154 & - & - & 8.8\\
		Unitree A1 & 13.7 & 33.5 & 21 & 3.3\\
		Unitree Aliengo & 22 & 35.3 & 20 & 1.5\\
		Unitree Go1 & 12 & 23.7 & 30 & 5\\
		Unitree Go2 & 15 & 23.7 & 30 & 5\\
		Unitree Go2-W & 18 & 23.7 & 30 & 2.5\\
		Unitree B2 & 60 & 200 & 23 & 6\\
		Unitree B2-W & 85 & 200 & 23 & 4.17\\
		Unitree As2 & 18 & 60 & 24 & 5\\
		Unitree A2 & 42 & 120 & 22 & 5\\
		DeepRobotics Lite3 & 12 & 24 & 26.2 & 5\\
		DeepRobotics Lynx M20 & 33 & 76.4 & 22.4 & 5\\
		DeepRobotics X30 & 56 & 150 & 16.1 & 4\\
		KAIST Hound\cite{shin_reinforcement_2025} & 45 & 50 & 85.26 & 6.5\\
		Boston Dynamics Spot\cite{miller_high-performance_2025} & 32.5 & 98 & 25.56 & 5.2\\
		KAIST Raibo\cite{kim_high-speed_2025} & 27.4 & 60 & 37.4 & 4\\
		KAIST Raibo2\cite{hwangbo_raibo2_2025} & 44.5 & 100 & 20.5 & 5.8\\
		Xiaomi CyberDog & 14 & 32 & 23 & 3.2\\
		Xiaomi CyberDog 2 & 8.9 & 12 & 31 & 1.6\\
		mjbots quad A1\cite{MjbotsQuadQuad} & 8 & 16 & 62.8 & 2.5\\
		Stanford Pupper\cite{kauStanfordPupperLowCost2022} & 2.1 & 1.8 & 60 & 0.8\\
		\hline
	\end{tabular}
\end{table}

\begin{table}[h] 
	\centering
	\caption{\textbf{Performance standards for China National Level athletes in the 100 m sprint.}
		The table summarizes the qualifying times for male and female athletes under the official athletics grading system.}
	\label{tab:100m_standard}

	\begin{tabular}{ccccc}
		\\
		\hline
		Category & Level & Event & Qualifying Time & Units\\
		\hline
		Male   & Level-1 & 100 m & 10.93          & s\\
		Male   & Level-2 & 100 m & \textbf{11.74} & s\\
		Male   & Level-3 & 100 m & 12.64          & s\\
		\hline
		Female & Level-1 & 100 m & 12.33          & s\\
		Female & Level-2 & 100 m & \textbf{13.04} & s\\
		Female & Level-3 & 100 m & 14.04          & s\\
		\hline
	\end{tabular}
\end{table}

\begin{table}[h]
\centering
\caption{\textbf{Typical physical parameters of the sagittal-plane model.} Values correspond to the nominal configuration.}
\label{tab:robot_params}
\begin{tabular}{cccc}
\hline
Symbols & Physical Meaning & Typical Values & Units \\
\hline
$m_0$ & Mass of torso & 6.15 & kg \\
$m_1$ & Mass of thigh & 2.2 & kg \\
$m_2$ & Mass of shank & 0.1 & kg \\
$m_3$ & Mass of toe & 0.5 & kg \\

$I_0$ & Inertia of torso & 0.10 & kg$\cdot$m$^2$ \\
$I_1$ & Inertia of thigh & 0.014 & kg$\cdot$m$^2$ \\
$I_2$ & Inertia of shank & $2\times10^{-5}$ & kg$\cdot$m$^2$ \\
$I_3$ & Inertia of toe & $1\times10^{-5}$ & kg$\cdot$m$^2$ \\

$L_0$ & Length of torso & 0.35 & m \\
$L_1$ & Length of thigh & 0.35 & m \\
$L_2$ & Length of shank & 0.10 & m \\
$L_3$ & Length of toe & 0.05 & m \\

$l_0$ & COM offset of torso & -0.10 & m \\
$l_1$ & COM offset of thigh & 0.024 & m \\
$l_2$ & COM offset of shank & -0.05 & m \\
$l_3$ & COM offset of toe & -0.02 & m \\

$r_1$ & Gear ratio of thigh & 8.5 & -- \\
$r_2$ & Gear ratio of knee & 8.5 & -- \\

$k_p$ & Spring stiffness & 6000 & N/m \\
$k_d$ & Spring damping & 10 & Ns/m \\

\hline
\end{tabular}
\end{table}

\begin{table}[h]
\centering
\caption{\textbf{Reward terms used for reinforcement learning controller training.}
The imitation stage emphasizes reference-motion guidance, whereas the relaxed stage continues training from the imitation policy with reweighted objectives for high-speed locomotion. Base coefficients are given as imitation/relaxed. Terms below the separator are added for the higher-speed BP2 training.}
\label{tab:rl_reward}
\begin{tabular}{p{0.24\linewidth} p{0.52\linewidth} p{0.14\linewidth}}
\hline
Reward term & Formulation & Base coefficient \\
\hline
Body height &
$r_h=c_h\exp[-80(z-z^{ref})^2]$ &
$0.05/0$ \\

Body attitude &
$r_R=c_R\exp[-80\|\mathbf{e}_z-\mathbf{e}^{ref}_z\|^2]$ &
$0.05/0$ \\

Joint position imitation &
$r_q=0.2c_m\exp[-2\|\mathbf{q}^{ref}-\mathbf{q}\|^2]$ &
$\bar c_J=0.6/0.1$ \\

Joint velocity imitation &
$r_{\dot q}=0.7c_m\exp[-0.02\|\dot{\mathbf{q}}^{ref}-\dot{\mathbf{q}}\|^2]$ &
$ $ \\

Velocity tracking &
$r_v=c_v[w\exp(-2\|\mathbf{v}-\mathbf{v}^{cmd}\|^2)+(1-w)\exp(-2\|\boldsymbol{\omega}-\boldsymbol{\omega}^{cmd}\|^2)]$ &
$\bar c_V=0.1/0.4$ \\

Torque regularization &
$r_\tau=c_\tau[0.9\exp(-0.5\|\boldsymbol{\tau}_n\|^2)+0.1\exp(-0.1\|\boldsymbol{\tau}_n-\boldsymbol{\tau}_{n,last}\|^2/\Delta t)]$ &
$0.1/0.2$ \\

Power regularization &
$r_P=\frac{c_P}{12}\sum_i \phi(P_i)$, where $\phi(P_i)=\exp(-0.01P_i)$ if $P_i>0$ and $\phi(P_i)=\exp(0.05P_i)$ otherwise &
$0/0.1$ \\

Action smoothness &
$r_s=c_s\exp[-3\|\mathbf{a}_t-\mathbf{a}_{t-1}\|_\infty]$ &
$0.1/0.2$ \\

Terminal penalty &
$r_T=-1$ if the robot enters a terminal state &
$-1/-1$ \\

\hline
\multicolumn{3}{c}{Additional reward term for higher-speed BP2 training} \\
\hline

Straight-running symmetry &
$r_{str}=\exp[-10\eta]\exp[-100((q^{abad}_{FR}+q^{abad}_{FL})^2+(q^{abad}_{RR}+q^{abad}_{RL})^2)]$ &
$0/0.1$ \\

\hline
\end{tabular}
\end{table}

\begin{table}[h]
\begin{center}
\begin{tabular}{p{0.22\linewidth} p{0.68\linewidth}}
\hline
Symbol & Description \\
\hline
$\mathbf{q}^{ref}$, $\dot{\mathbf{q}}^{ref}$ &
Reference joint position and velocity generated by the kinematic gait generator. \\

$\mathbf{q}$, $\dot{\mathbf{q}}$ &
Measured joint position and velocity of the robot. \\

$\mathbf{v}^{cmd}$, $\boldsymbol{\omega}^{cmd}$ &
Commanded linear and angular velocities. \\

$\mathbf{v}$, $\boldsymbol{\omega}$ &
Measured body linear and angular velocities. \\

$\boldsymbol{\tau}_n$ &
Normalized joint torque. \\

$\boldsymbol{\tau}_{n,last}$ &
Normalized joint torque at the previous control step. \\

$P_i$ &
Mechanical power of the $i_{th}$ actuator. \\

$\mathbf{a}_t$, $\mathbf{a}_{t-1}$ &
Policy actions at the current and previous control steps. \\

$q^{abad}_{FR}$, $q^{abad}_{FL}$, $q^{abad}_{RR}$, $q^{abad}_{RL}$ &
Abduction/adduction joint angles of the front-right, front-left, rear-right, and rear-left legs. \\

$\eta$ &
Activation factor for the straight-running symmetry reward, defined as $\eta=\max(|v_y^{cmd}|/v_y^{max},|\omega_z^{cmd}|/\omega_z^{max})$. \\

$w$ &
Weight balancing linear and angular velocity tracking, defined as $w=\min[(v_x^{max}+v_y^{max})/(v_x^{max}+v_y^{max}+\omega_z^{max}),0.6]$. \\

$z$, $z^{ref}$ &
Measured body height and reference body height. \\

$\mathbf{e}_z$, $\mathbf{e}^{ref}_z$ &
Measured and reference body vertical axes. \\

$\beta$ &
Speed-dependent relaxation factor, defined as $\beta=(|v_x^{cmd}|/v_x^{max})^{0.3}$. \\

$\bar c_J$ &
Base coefficient for reference-motion imitation reward. \\

$\bar c_V$ &
Base coefficient for velocity-tracking reward. \\

$c_m$ &
Speed-dependent coefficient for reference-motion imitation, defined as $c_m=c_J+c_V-c_v$. \\

$c_v$ &
Speed-dependent coefficient for velocity tracking, defined as $c_v=\beta c_V+(1-\beta)c_J$. \\

$\Delta t$ &
Control time step. \\
\hline
\end{tabular}
\end{center}
\end{table}


\clearpage 



\end{document}